\documentclass[10pt,twocolumn,letterpaper]{article}

\usepackage[pagenumbers]{cvpr} %

\usepackage{xcolor}

\usepackage{colortbl}

\usepackage{multirow}
\usepackage{verbatim}
\usepackage{bm}

\colorlet{best3}{white}
\colorlet{best2}{yellow!20!white}
\colorlet{best1}{green!20!white}

\newcommand{\minihead}[1]{\vspace{0.5em}\noindent\textbf{#1}}

\usepackage{tikz}
\newcommand{\coloredCell}[3]{\definecolor{mycolor}{HTML}{#2}\tikz[baseline=(char.base)]{\node[fill=mycolor,inner ysep=2pt, inner xsep=5pt, minimum width=#1, text width=#1, align=right] (char) {#3};}}

\newcommand{\cellCD}[3][24pt]{\coloredCell{#1}{#2}{#3}}

\newcommand{\tablestyle}[2]{
  \centering
  \setlength{\tabcolsep}{#1}
  \renewcommand{\arraystretch}{#2}
  \footnotesize
}

\usepackage{array}
\newcolumntype{x}[1]{>{\centering\arraybackslash}p{#1pt}}
\newcolumntype{y}[1]{>{\raggedright\arraybackslash}p{#1pt}}
\newcolumntype{z}[1]{>{\raggedleft\arraybackslash}p{#1pt}}

\usepackage{dcolumn}
\newcolumntype{d}{D{:}{\,:}{-1}}

\usepackage{siunitx}

\newcommand{\alg}{\textsc{FastMap}\xspace}

\makeatletter
\renewcommand{\paragraph}{%
  \@startsection{paragraph}{4}%
  {\z@}{0.50ex \@plus 1ex \@minus .2ex}{-1em}%
  {\normalfont\normalsize\bfseries}%
}
\makeatother

\newcommand{\mA}{\mathbf{A}}

\newcommand{\mE}{\mathbf{E}}
\newcommand{\mF}{\mathbf{F}}

\newcommand{\mH}{\mathbf{H}}

\newcommand{\mR}{\mathbf{R}}

\newcommand{\mW}{\mathbf{W}}

\newcommand{\ve}{\bm{e}}

\newcommand{\vo}{\bm{o}}

\newcommand{\vt}{\bm{t}}

\newcommand{\vw}{\bm{w}}
\newcommand{\vx}{\bm{x}}

\definecolor{cvprblue}{rgb}{0.21,0.49,0.74}
\usepackage[pagebackref,breaklinks,colorlinks,citecolor=cvprblue]{hyperref}

\def\paperID{310} %
\def\confName{3DV\xspace}
\def\confYear{2026\xspace}

\title{\alg: Revisiting Structure from Motion through First-Order Optimization}

\author{
    Jiahao Li\textsuperscript{1}\thanks{core contributors} \quad
    Haochen Wang\textsuperscript{1}\footnotemark[1] \quad
    Muhammad Zubair Irshad\textsuperscript{2} \quad
    Igor Vasiljevic\textsuperscript{2} \\
    Matthew R.\ Walter\textsuperscript{1} \quad
    Vitor Campagnolo Guizilini\textsuperscript{2}\thanks{joint supervision \& equal contributions} \quad
    Greg Shakhnarovich\textsuperscript{1}\footnotemark[2] \\
    \textsuperscript{1}TTI-Chicago \quad\quad
    \textsuperscript{2}Toyota Research Institute \\
    {\tt\small \{jiahao,whc,mwalter,greg\}@ttic.edu} \\
    {\tt\small \{zubair.irshad,igor.vasiljevic,vitor.guizilini\}@tri.global}
}

\begin{document}
\maketitle
\begin{abstract}
We propose \alg, a new global structure from motion method focused on speed and simplicity. Previous methods like COLMAP and GLOMAP are able to estimate high-precision camera poses, but suffer from poor scalability when the number of matched keypoint pairs becomes large, mainly due to the time-consuming process of second-order Gauss-Newton optimization. Instead, we design our method solely based on first-order optimizers. To obtain maximal speedup, we identify and eliminate two key performance bottlenecks: computational complexity and the kernel implementation of each optimization step. Through extensive experiments, we show that \alg is up to $10\times$ faster than COLMAP and GLOMAP with GPU acceleration and achieves comparable pose accuracy. Project webpage: \url{https://jiahao.ai/fastmap}.
\end{abstract}
    
\section{Introduction}

\label{sec:intro}

Data is the fuel for state-of-the-art computer vision systems. Recently, synthetic 3D datasets~\cite{tartanair,dsetspring,pointodyssey,greff2022kubric,urbansyn,deitke2023objaverse}
have been scaled up to provide supervision for diverse tasks such as
Visual-SLAM~\cite{teed2021droid}, 3D point tracking~\cite{karaev2024cotracker,harley2025alltracker}, 
3D asset generation~\cite{shi2023mvdream,li2023instant3d,trellis}, etc. 
However, scaling up real-world 3D data remains difficult due to the lack of ground-truth camera poses. Many applications such as monocular depth estimation~\cite{ranftl2020towards, bhat2023zoedepth, ke2024repurposing} and learning-based 3D reconstruction~\cite{wang2024dust3r, duisterhof2024mast3r, wang2025vggt} still rely on pseudo-ground-truth produced by pure geometry-based structure from motion 
systems (SfM)
such as COLMAP~\cite{schoenberger2016sfm}. However, COLMAP is slow---processing a scene consisting of thousands of images can take multiple days. Global SfM methods such as GLOMAP~\cite{pan2024glomap} improve upon COLMAP's speed, but still take many hours to converge on large scenes.  Efficiently scaling learning-based systems to more training data requires a fast and high-quality ground-truth annotator.

\begin{figure}[!t] %
\centering
\includegraphics[width=1.0\columnwidth]{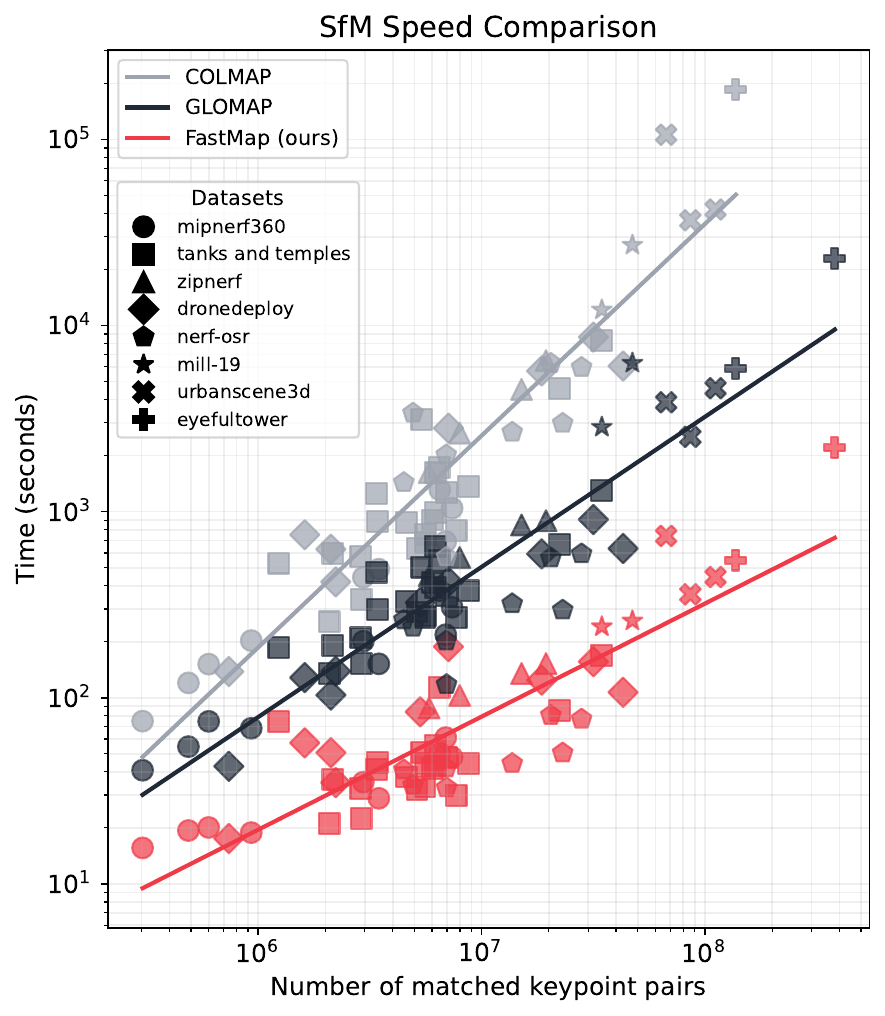}
\caption{
    Timing of \textcolor{red}{\alg} compared to \textcolor{gray}{COLMAP} %
    and GLOMAP~%
    (all with GPU acceleration on a single A6000) on scenes from eight datasets, %
    excluding the matching stage for all methods. Note the \textbf{logarithmic} time scale. Lines represent a least squares power function fit to timing across multiple datasets, as a function of the number of matched keypoint pairs.
}
\label{fig:timing}
\vspace{-4mm}
\end{figure}
A typical SfM pipeline spends most of the time doing optimization. For example, COLMAP~\cite{schoenberger2016sfm} registers one image at a time, and performs a round of global bundle adjustment optimization every few images to avoid drifting. GLOMAP~\cite{pan2024glomap} estimates global translation by optimizing the camera centers and 3D points from random initializations, and uses a final round of bundle adjustment at the end for refinement. These optimization problems are nonlinear~\cite{triggs2000bundle} and require iterative methods to solve. Specifically, quasi second-order methods such as Levenberg-Marquardt (LM)~\cite{Agarwal_Ceres_Solver_2022} are the standard choice. 
LM is a trust-region variant of the Gauss-Newton method, which estimates the Hessian with the residual Jacobian. While several techniques have been adopted to speed up LM optimization, such as the Schur complement trick and sparse Cholesky decomposition, each iteration still takes long wall-clock time to compute. %
In contrast, adaptive first-order methods~\cite{kingma2014adam}
can potentially eliminate the scalability bottleneck and simplify algorithm design, but are not well explored in the current SfM literature. 

To bridge this gap, we propose \alg, a global SfM method that relies only on first-order optimization. We identify and tackle two main speed bottlenecks when switching from second-order to first-order optimizers. First, many sub-problems in SfM, such as bundle adjustment~\cite{triggs2000bundle} and global positioning~\cite{pan2024glomap}, jointly optimize camera poses and 3D points. Usually, the number of 3D points is orders-of-magnitude larger than the number of image pairs. This is not the main issue for second-order methods because they spend most of the time solving the reduced linear system from the Schur complement, but can be a significant bottleneck for first-order methods that only require computing gradients. To address this, we design our method such that all the optimization problems involved have a per-step computational complexity independent of the number of 3D points. 

The second bottleneck comes from the implementation. While it is straightforward to implement gradient descent 
with modern deep learning Autograd engines~\cite{paszke2019pytorch}, we find that it leads to sub-optimal utilization of GPU resources. This is mainly due to the large overhead of kernel launching, unnecessary data movement between global and shared memory,
and improper kernel choice by the library. Instead, we use kernel fusion to perform 
forward and backward steps
in one CUDA kernel, which significantly speeds up each optimization step. 

Extensive experiments on 8 different datasets demonstrate that our method can be up to $10\times$ faster than both COLMAP and GLOMAP with GPU-accelerated Ceres solver~\cite{Agarwal_Ceres_Solver_2022} (\cref{fig:timing}). It also achieves comparable accuracy to these two state-of-the-art methods in terms of both pose accuracy and novel view synthesis quality. 

In summary, we introduce \alg, a new SfM framework with the following contributions:
\begin{itemize}
    \item We show that first-order optimization can be used to make a scalable and accurate SfM system.
    \item We design a fully 3D point-free pipeline that is friendly to first-order optimizers.
    \item We show that kernel fusion can significantly speed up gradient computation by eliminating overhead.
\end{itemize}

\section{Related Work}
\label{sec:related}

\minihead{Global SfM Systems} \emph{Incremental} SfM methods like COLMAP~\cite{schoenberger2016sfm} are state-of-the-art in accuracy and robustness, but \emph{global} SfM systems are catching up~\cite{pan2024glomap}.  These methods solve for all camera poses at once to avoid registering images sequentially, dramatically improving run-time.
OpenMVG~\cite{moulon2017openmvg} and Theia~\cite{sweeney2015theia} are two popular global SfM systems which are fast, but trail COLMAP in accuracy and robustness~\cite{pan2024glomap}. %
HSfM~\cite{cui2017hsfm} is a hybrid approach that combines incremental and global approaches, estimating rotations globally and translations incrementally. 
 Unlike prior global approaches that first perform translation averaging and then triangulation, GLOMAP~\cite{pan2024glomap} combines both steps, %
 solving for camera translations and 3D points in one global step.  They report results on par with COLMAP, but with large speed improvements both on small-~\cite{schops2017multi} and large-scale~\cite{sarlin2022lamar} datasets.

\minihead{Global Rotation and Translation} In a typical global SfM system, global rotation and translation are estimated directly from pairwise relative motions.  \citet{hartley2013rotation} provides a good tutorial on rotation averaging. \citet{govindu2001combining} frames motion estimation as a global optimization problem, and \citet{martinec2007robust} first solves for camera rotations using pairwise constraints and then obtains translations from a linear system using epipolar constraints. \citet{wilson2020distribution} 
propose a more stable optimizer for rotation averaging. 
Many existing approaches struggle when baseline lengths differ significantly~\cite{pan2024glomap}.
LUD~\cite{ozyesil2015robust} and \citet{zhuang2018baseline} focus on improving stability and robustness in ill-conditioned scenarios. 
\citet{jiang2013global} introduces a linearized approach that enforces constraints across camera triplets, ensuring consistency in multi-view configurations. \citet{wilson2014robust} propose improving translation estimation by combining outlier removal with a simplified solver.

\noindent\textbf{Learning-based SfM}
Learning-based SfM methods vary in the degree of their departure from the traditional SfM pipeline and in their tradeoff between speed and accuracy.
VGGSfM~\cite{wang2024vggsfm} hews closely to the traditional SfM methodology, building on point-tracking methods to propose a fully differentiable SfM method that includes bundle adjustment.
Flowmap~\cite{smith2024flowmap} uses pretrained optical flow and point tracking networks and a depth CNN to optimize per-scene global poses, calibration, and depth maps using gradient descent.
Ace-Zero~\cite{brachmann2024scene} uses a trained dense 3D scene coordinate regressor as an alternative to triangulation and registration in incremental SfM, instead incrementally relocalizing with the learned coordinate regressor. The DUSt3R~\cite{wang2024dust3r} architecture, which maps image pairs to \textit{point maps}, initiated a new paradigm in learned SfM.  
DUSt3R pointmap estimates from image pairs can be used for camera calibration, depth estimation, correspondence, pose estimation and dense reconstruction. Many recent works~\cite{yang2025fast3r, lu2025matrix3d, wang2025vggt} improve upon DUSt3R in various ways, such as efficiently processing more input images~\cite{yang2025fast3r}, using diffusion models~\cite{lu2025matrix3d}, and predicting more 3D geometry attributes~\cite{wang2025vggt}. In particular, MASt3R~\cite{leroy2024grounding} upgrades DUSt3R using dense correspondences from the predicted DUSt3R pointmap pairs, and MASt3R-SfM~\cite{duisterhof2024mast3r} incorporates a global alignment stage, offering a full-fledged SfM system based on MASt3R.

\section{Method}
\label{sec:method}

\begin{figure}[!t] %
    \centering
    \includegraphics[width=1.0\columnwidth]{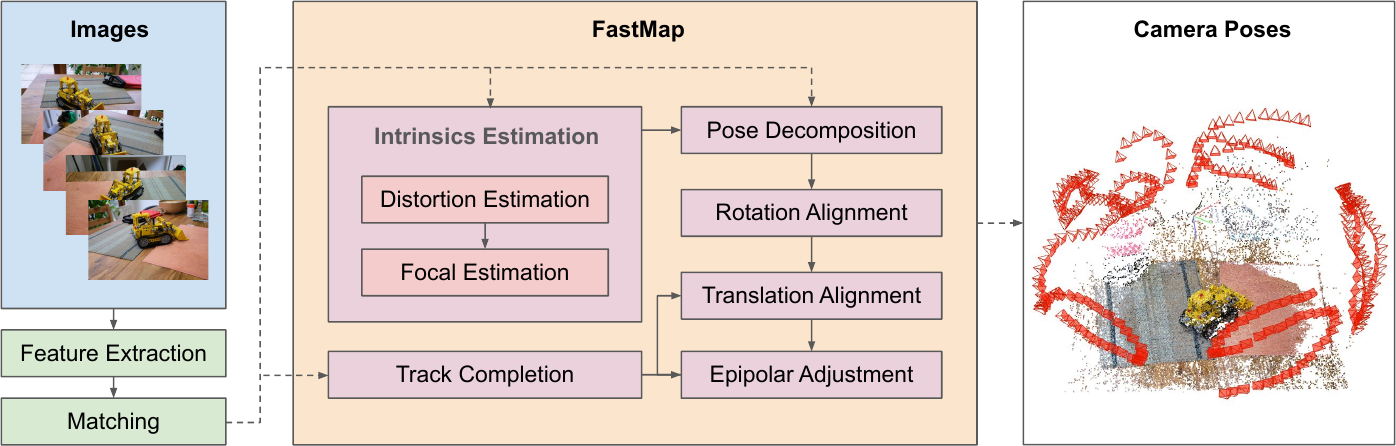}
    \caption{An overview of \alg. Input images are processed using feature extraction and matching. Given the matching results, \alg estimates the intrinsics and extrinsics of the cameras. Finally a sparse point cloud is generated by triangulation.} %
    \label{fig:pipeline}
\vspace{-3mm}
\end{figure}

\minihead{Overview} 
Our proposed \alg framework (\cref{fig:pipeline}) consists of multiple stages roughly in sequential order. In this section we introduce the algorithmic details. \cref{sec:method-intrinsics} describes how we estimate the distortion parameters and focal lengths for each camera after extracting and matching keypoints. Then, \cref{sec:method-first-order-optimization} analyzes the pros and cons of first-order vs second-order optimization, and \cref{sec:method-optimization-problems} describes all the optimization problems for global pose estimation and refinement. Finally, we discuss kernel fusion (\cref{sec:method-kernel-fusion}) for further speed-up.

\minihead{Matching} \alg's matching stage is identical to that of both COLMAP and GLOMAP: it involves first extracting and matching keypoints from the input images, followed by geometric verification of the resulting image pairs~\cite{schoenberger2016sfm}. The output of the matching stage consists of the 
set of inlier keypoint pairs and either an estimated fundamental matrix $\mF_{ij}$ or a homography matrix $\mH_{ij}$ (the latter if it is consistent with sufficiently many inlier matches) for each image pair $(i,j)$ with enough correspondences.

\subsection{Intrinsics Estimation}
\label{sec:method-intrinsics}

Accurate intrinsics estimation has a direct impact on the precision of relative pose decomposition, which is critical for later stages. In this section, we describe the algorithms that \alg employs to estimate the focal lengths and distortion from the matching results. 

\minihead{Camera Assumptions} We use a one-parameter radial distortion model. The principal point is fixed to be the center, therefore the only intrinsics parameters to estimate are the distortion parameter and the focal length. We also assume that all the images are taken with a small number of distinct cameras, and that we know which images are from the same camera. In practice, this can be inferred from image resolutions, EXIF tags, file and directory names, etc.

\minihead{Distortion Estimation} 
We formulate distortion estimation as the problem of finding the distortion parameters that result in the most consistent two-view geometric model for each image pair (e.g., the fundamental matrix estimated from undistorted keypoints has the lowest epipolar error). We do so using the one-parameter division undistortion model~\cite{fitzgibbon2001simultaneous, barreto2005fundamental, erdnuss2021review, henrique2013radial}
\begin{equation}
    x_u = \frac{x_d}{1+\alpha r_d^2} \qquad
    y_u = \frac{y_d}{1+\alpha r_d^2},
\end{equation}
where $(x_d, y_d)$ and $(x_u, y_u)$ are the distorted and undistorted coordinates, respectively, $r_d = \sqrt{x_d^2+y_d^2}$ and $r_u = \sqrt{x_u^2+y_u^2}$, and $\alpha$ is the distortion model parameter. The model can be inverted in closed form to apply distortion to keypoints~\cite{erdnuss2021review}. %
We found this model to be more convenient than the commonly used, but difficult to invert Brown-Conrady distortion model~\cite{duane1971close}.

We use brute-force interval search to estimate the distortion parameter $\alpha$. Given a set of image pairs that share the same $\alpha$, we sample set of candidate values from an interval $[\alpha_{\text{min}}, \alpha_{\text{max}}]$, and evaluate the average epipolar errors for each candidate after undistorting and re-estimating the fundamental matrices based on the sampled $\alpha$ (we ignore all the homography matrices at this stage). This method directly minimizes our objective (epipolar error) and takes into account information from multiple different image pairs, improving robustness to noise. Moreover, each candidate can be evaluated independently, making it highly parallelizable on a GPU. In the supplementary material (\cref{app:distortion}), we provide more details regarding how to accelerate the above method with hierarchical sampling, and discuss generalizations to images with different distortion parameters.

\minihead{Focal Length Estimation} 
We use the estimated distortion model to undistort all the keypoints, and re-estimate the fundamental and homography matrices. At this point, the only remaining unknown intrinsic parameter for each camera is the focal length.
We estimate the focal length based on the re-estimated fundamental matrices from undistorted keypoints. While this is a well studied problem~\cite{hartley1993extraction, kanatani2000closed, sweeney2015optimizing, barath2017minimal, kocur2024robust}, existing methods are
susceptible to noise or require nonlinear optimization. Instead, we employ an interval search strategy similar to that used for distortion estimation, but with a different objective.

Our method is based on the well-known property that a $3 \times 3$ matrix is an essential matrix if and only if its singular values are such that $\lambda_1=\lambda_2$ and $\lambda_3=0$ (where $\lambda_1 \geq \lambda_2 \geq \lambda_3$)~\cite{faugeras1993three, hartley2003multiple}. %
Given the correct fundamental matrix $\mathbf{F}$ and intrinsics matrix $\mathbf{K}$, the essential matrix $\mathbf{E}=\mathbf{K}^\top \mathbf{F} \mathbf{K}$ should satisfy $\frac{\lambda_1}{\lambda_2}=1$. If all images share the same intrinsics, with a set of fundamental matrices $\{\mathbf{F}_i\}$, we can evaluate the accuracy of a candidate focal length $f$ using the singular value ratio above. Letting $\lambda_1^{(i)} \geq \lambda_2^{(i)} \geq \lambda_3^{(i)}$ be the singular values of $\mathbf{K}^\top \mathbf{F}_i \mathbf{K}$, where $\mathbf{K}$ is a function of $f$, we can measure the validity of $f$ as
\begin{align} \label{eq:focal-vote}
    v = \sum_{i} \exp\left(\frac{1-\lambda_1^{(i)}/\lambda_2^{(i)}}{\tau}\right),
\end{align}
where $\tau$ is a temperature hyper-parameter. %
Intuitively, the above formula is close to one when $\lambda_1^{(i)}/\lambda_2^{(i)}$ is close to one, and decreases exponentially as $\lambda_1^{(i)}/\lambda_2^{(i)}$ increases.

We sample a set of candidate focal lengths and evaluate them using Eqn.~\ref{eq:focal-vote}. We choose the candidate with the highest value~\eqref{eq:focal-vote} as the final estimate. This method can be easily generalized to images with different focal lengths (see \cref{app:focal} in the supplementary for details).
After estimating the focal length, we transform the keypoints, fundamental matrices, and homography matrices using the estimated intrinsic matrices so all components are calibrated.

\subsection{First-order Optimization}
\label{sec:method-first-order-optimization}

\minihead{Levenberg-Marquardt} Most of the previous SfM methods use quasi second-order methods such as Levenberg-Marquardt (LM) for optimization. Almost all methods use LM for bundle adjustment~\cite{triggs2000bundle}, and GLOMAP~\cite{pan2024glomap} uses it for global translation alignment. Levenberg-Marquardt is a Gauss-Newton method that first approximates the Hessian with Jacobians of the residuals, and then solves a large linear system to compute the update direction. Techniques like the Schur's complement and sparse Cholesky decomposition are used to improve computational efficiency by exploiting the sparsity of the system~\cite{triggs2000bundle, Agarwal_Ceres_Solver_2022}.

While second-order methods can converge quickly near the optimum, they suffer from poor scalability. Even with the Schur complement method, each step requires solving a reduced linear system whose size grows quadratically with the number of images. In practice, this results in a cubic-time cost in the number of images, which dominates the computation of the update direction. In very large and densely-connected problems, this becomes prohibitively expensive, even with GPU acceleration. To address this, many frameworks employ the preconditioned conjugate gradient (PCG) method, which approximately solves the reduced linear system at a per-iteration cost that scales only quadratically in the number of images. However, PCG slows convergence and introduces implementation complexity.

\minihead{First-order Optimization} On the other hand, first-order optimization methods are prevalent in other fields of computer vision, thanks to the success of deep learning. Optimizing a neural network with a large number of parameters is only tractable with first-order methods, and many adaptive gradient methods exist that  accelerate the naive gradient descent.
In this paper, we investigate the use of first-order optimization methods in SfM.

\minihead{Efficiency} Unfortunately, first-order methods usually converge much more slowly than Gauss-Newton methods in terms of the decrease in loss at each iteration (i.e., they require more iterations). The key to the success of using first-order optimization in SfM is to make the computation of each step as fast as possible. We identify the two most important speed bottlenecks:
\begin{enumerate}
    \item \textit{3D points}:
    One of the most important components of a typical SfM pipeline is bundle adjustment~\cite{triggs2000bundle}, which jointly optimizes camera poses and 3D points. In practice, the number of 3D points is usually orders-of-magnitude larger than the number of image pairs. To make Gauss-Newton tractable in this setting, methods employ the Schur complement method~\cite{triggs2000bundle} to first eliminate the 3D point variables to form a reduced system independent of the number of points, and then reciver the 3D points via back-substitution. In this stage, solving the reduced system usually dominates the compute time. However, if we switch to gradient descent, the main bottleneck becomes computing the forward and backward passes for the 3D point parameters. To address this, we design all the optimization problems in our method (\cref{sec:method-optimization-problems}) so that at each iteration, the computation complexity is independent of the number of 3D points.
    \item \textit{Kernel implementation}:
    The optimization problems that \alg solves can be easily implemented using modern Autograd frameworks such as PyTorch. These libraries are highly optimized for large-scale deep learning applications that involve a lot of linear operations on large tensors. In our case, most of the operations are relatively small (e.g., $3\times 3$ matrix multiplication), and naively implementing everything with high-level PyTorch optimizations induces significant kernel launching and data movement overhead. We solve this problem through kernel fusion (\cref{sec:method-kernel-fusion}), which eliminates most of the overhead and increases GPU utilization.
\end{enumerate}
\vspace{0.5em}\noindent  Section~\ref{sec:method-optimization-problems} introduces all the optimization problems present in our method. They are chosen such that the computational complexity of each step is independent of the number of 3D points. In \cref{sec:method-kernel-fusion}, we describe the kernel fusion technique to fully exploit the power of GPUs for even more speedup.

\subsection{Optimization Formulations}
\label{sec:method-optimization-problems}
Here, we introduce the optimization-based formulations of estimating and finetuning the global poses.

\minihead{Global Rotation} 
With the estimated intrinsics, we can decompose the fundamental and homography matrices into relative rotations and translations~\cite{hartley2003multiple, malis2007deeper}. Given the set of image pairs $\mathcal{P}=\{(i, j)\}$ and the corresponding relative rotation matrices $\{\mR^{i\rightarrow j}\}_{(i, j)\in \mathcal{P}}$, \alg next estimates the world-to-camera global rotation $\mathbf{R}^{(i)}$ matrix for each image $i$. We formulate this as an optimization of a loss defined over all image pairs $\mathcal{P}=\{(i, j)\}$ %
\begin{equation} \label{eq:rotation-objective}
    \mathcal{L}_{\text{R}} = \frac{1}{|\mathcal{P}|}\sum_{(i, j)\in\mathcal{P}} d( \mR^{(j)}, \mR^{i\rightarrow j} \mR^{(i)}),
\end{equation}
where $d(\cdot, \cdot)$ is the geodesic distance between rotations %
\begin{equation}
    d(\mR, \mR^\prime) = \cos^{-1} \left( \frac{\text{Tr}(\mR^T \mR^\prime) - 1}{2} \right).
\end{equation}
For simplicity, we parameterize the global rotation matrices $\mathbf{R}_i$ using a differentiable 6D representation~\cite{zhou2019continuity}.

Unfortunately, directly optimizing the above objective from random initialization of $\mathbf{R}_i$ is prone to local minima. We use a slightly modified version of the method proposed by \citet{martinec2007robust} to obtain a good initialization. The basic idea of the method is that although the column vectors in a rotation matrix are constrained by orthogonality, each column vector alone is only subject to a unit length constraint. If we consider one column at a time, we can formulate the optimization as as a least squares problem and solve it using SVD. See the supplementary material (\cref{app:rotation-initialization}) for details of this initialization scheme.

\minihead{Global Translation} 
After global rotation alignment, we re-estimate the relative translations between image pairs (\cref{app:translation}). The next step is to utilize these relative translations to estimate the 3D coordinates of the camera centers in a common (world) coordinate frame, up to a similarity transformation. This step is usually called \textit{translation averaging}. It is notoriously susceptible to noise, and making it robust to all kinds of scenarios is the focus of many global SfM papers~\cite{jiang2013global, wilson2014robust, ozyesil2015robust, zhuang2018baseline, pan2024glomap}. However, this is not the focus of our paper and so we choose a relatively simple method to tackle this problem, which we find to be sufficient for most of the scenes we evaluate on. A more robust design for this stage is left for future work.

Given %
world-to-camera rotations $\{\mR_i\}_{1\leq i\leq N}$ for $N$ images and unit-length relative translations $\{\vt^{i\rightarrow j}\}_{(i,j)\in\mathcal{P}}$ for a set of image pairs $\mathcal{P}$, we compute the normalized vector from the camera centers of image $i$ to $j$ in world coordinates
\begin{equation}
\vo^{i\rightarrow j} = -\mR_j^\top \vt^{i\rightarrow j}.
\end{equation}

We estimate the camera locations $\{\vo_i\}_{1\leq i\leq N}$ in the world frame by minimizing the error between the normalized relative translation $\frac{\vo_j - \vo_i}{\|\vo_j - \vo_i\|_2}$ and the target $\vo^{i\rightarrow j}$ above with gradient descent 
\begin{equation}\label{eq:global-translation-objective}
    \mathcal{L}_{\text{t}} = \frac{1}{|\mathcal{P}|} \sum_{(i,j)\in\mathcal{P}} \left\| \frac{\vo_j - \vo_i}{\|\vo_j - \vo_i\|_2} - \vo^{i\rightarrow j} \right\|_1.
\end{equation}
Unlike global rotation optimization, this objective can often be effectively optimized from a random initialization. %
GLOMAP~\cite{pan2024glomap} makes a similar observation, but it optimizes poses and 3D points jointly, and is much more computationally expensive.

Although random initialization works surprisingly well for the objective in Eqn.~\ref{eq:global-translation-objective}, it occasionally produces a small number of outliers. To deal with this, we perform multiple independent runs from different random initializations and merge the solutions as the initialization for the final optimization loop. Please see \cref{app:translation-initializations} for details.

\minihead{Epipolar Adjustment} 
A typical SfM pipeline relies on \emph{bundle adjustment} (BA)~\cite{triggs2000bundle} to jointly refine the camera poses and inferred 3D points. Directly implementing BA with first-order optimizers is computationally expensive when the number of points is large. 
Instead, we refine the poses from previous stages using \emph{re-weighting epipolar adjustment}, an optimization method~\citep{rodriguez2011reduced} for which the computational complexity in each iteration is linear only in the number of image pairs, not in the number of points.

In relative translation re-estimation, we obtain a set of image pairs with number of inliers above some threshold. We denote the set of such image pairs as $\mathcal{P}=\{(i_n, j_n)\}_{1\leq n\leq |\mathcal{P}|}$ (abusing the notation for the original set of images above), where $i_n$ and $j_n$ are the indices of the first and second images in the pair. %
For an image pair $(i_n, j_n)\in\mathcal{P}$, we represent the set of point pairs as $\mathcal{Q}_n=\{(\vx^{(1)}_{nm}, \vx^{(2)}_{nm})\in\mathbb{R}^2\times\mathbb{R}^2\}_{1\leq m\leq |\mathcal{Q}_n|}$, and let  $\tilde{\mathcal{Q}}_n=\{(\tilde{\vx}^{(1)}_{nm}, \tilde{\vx}^{(2)}_{nm})\in\mathbb{P}^2\times\mathbb{P}^2\}_{1\leq m\leq |\tilde{\mathcal{Q}}_n|}$ be the set of point pairs in normalized homogeneous coordinates.

Using estimated initializations from the previous stages, we optimize over the world-to-camera global rotations and translations to minimize the absolute epipolar error
\begin{equation} \label{eq:epipolar-objective}
    \mathcal{L}_{\text{e}} = \frac{1}{Z} \sum_{n=1}^{|\mathcal{P}|}\sum_{m=1}^{|\tilde{\mathcal{Q}}_n|} \lvert \tilde{\vx}^{(2)\top}_{nm}\mE_n\tilde{\vx}^{(1)}_{nm} \rvert,
\end{equation}
where $Z=\sum_{n=1}^{|\mathcal{P}|}|\tilde{\mathcal{Q}}_n|$ is the total number of point pairs, and $\mE_n$ is the essential matrix computed from the global rotations and translations for images $i_n$ and $j_n$.

Evaluating Eqn.~\ref{eq:epipolar-objective} for every iteration is expensive because it involves every point pair. However, if we replace the cost terms with the L2 loss (as in \citet{rodriguez2011gea}), the overall objective can be re-organized to aggregate terms that involve point pairs shared by the same image pair into a compact quadratic form (see \cref{app:epipolar})%
\begin{equation} \label{eq:epipolar-objective-l2}
    \mathcal{L}_{\text{e}}\hspace{-.2em}=\hspace{-.2em} \frac{1}{Z}\sum_{n=1}^{|\mathcal{P}|}\sum_{m=1}^{|\tilde{\mathcal{Q}}_n|} ( \tilde{\vx}^{(2)\top}_{nm}\mE_n\tilde{\vx}^{(1)}_{nm} )^2=\frac{2}{Z}\sum_{n=1}^{|\mathcal{P}|} \ve_n^\top \mW_n \ve_n,
\end{equation}
where  $\ve_n=\text{flatten}(\mE_n)\in\mathbb{R}^9$, and $\mW_n\in\mathbb{R}^{9\times 9}$ is a matrix computed from all the point pairs in $\tilde{\mathcal{Q}}_n$. Note that only $\ve_n$ is a function of the parameters to be optimized. The matrices $\mW_n$ can be pre-computed for each image pair before optimization. With the precomputed $\mW_n$, the cost of each optimization step is linear in the number of image pairs. 

The expedience of Eqn.~\ref{eq:epipolar-objective-l2} comes with the side effect that the L2 loss is sensitive to outliers. We propose to robustify the loss function
but still preserve the benefit of pre-computation with \emph{iterative re-weighted least squares (IRLS)}. The intuition is that if we have an initialization close to the optimum, the L1 loss, which is more robust to outliers, can be approximated by a weighted L2 loss. In other words, for some differentiable computed scalar value $z$, we have $ z^2\approx z^2 / |\hat{z}|$, where $\hat{z}$ is the value of $z$ at initialization. In our case, after global translation alignment, we already have a good initialization of global poses, so we can compute the absolute epipolar error $|\hat{\epsilon}_{nm}|$ for each point pair, and use it to weight the L2 loss used above to get an approximate robust L1 loss
\begin{equation}
        \hat{\mathcal{L}}_{\text{e}}
        \hspace{-.3em}= \frac{1}{Z}\sum_{n=1}^{|\mathcal{P}|}\sum_{m=1}^{|\tilde{\mathcal{Q}}_n|} \frac{( \tilde{\vx}^{(2)\top}_{nm}\mE_n\tilde{\vx}^{(1)}_{nm} )^2}{|\hat{\epsilon}_{nm}|}
        = \frac{2}{Z}\sum_{n=1}^{|\mathcal{P}|} \ve_{n}^\top \hat{\mW}_n \ve_{n},
\end{equation}
where $\hat{\mW}_n$ is similar to $\mW_n$ in Eqn.~\ref{eq:epipolar-objective-l2}, but computed from $\tilde{\mathcal{Q}}_n$ weighted by $|\hat{\epsilon}_{nm}|$.

After a round of optimization, we can better approximate the L1 loss by re-computing the weights and then doing another optimization loop.
To further reduce the impact of outliers, we periodically filter out point pairs with large epipolar error. We start from a relatively large initial threshold and gradually decrease it to a pre-determined minimum.

The above optimization problem only involves camera poses. We can also optimize the focal lengths by incorporating them into the computation of the essential matrices (in this case, the results are actually fundamental matrices).

\subsection{Kernel Fusion}
\label{sec:method-kernel-fusion}

A standard way to implement gradient descent in the above algorithms is to use the Autograd  feature in modern deep learning libraries such as PyTorch~\cite{paszke2019pytorch}. However, the tensors in our setting are mostly batches of small matrices and vectors of shapes $B\times 3\times 3$ and $B\times 3$, where $B$ is the number of image pairs, and a naive PyTorch implementation introduces some significant bottlenecks:
\begin{enumerate}
    \item \textit{Kernel Launching Overhead:} When the scene is relatively small (i.e., the number of image pairs is small), the kernel launching overhead dominates the running time and is limited by the CPU speed.
    \item \textit{Data Movement:} Computing the objective and gradients involves a series of small operations (e.g., $3\times 3$ matrix multiplication and cross product). Each operation involves reading the data from the high-latency global memory to the fast on-chip shared memory and writing back when finished. his leads to substantial inefficiency and makes the computation predominantly memory-bound.
    \item \textit{Kernel Design:} PyTorch kernels are optimized for deep learning workload, which usually consists of linear operations on large tensors. These kernels can lead to sub-optimal performance when applied to tensors without the assumed shapes.
\end{enumerate}

\noindent To address this problem, we fuse all the operations for computing the gradients, including forward and backward passes, into one single custom CUDA kernel. This introduces challenges in shared-memory management when the computation involves many small operations. However, since almost all input and intermediate tensors have one of the three shapes ($B\times 3\times 3$, $B\times 3$, or simply $B$), we can efficiently reuse shared-memory slots to limit the reduction in thread occupancy. In \cref{tab:kernel_fusion} we show that the fused kernel can be more than two orders-of-magnitude faster than a naive PyTorch implementation under different scene sizes and hardware settings. Please refer to the ablation study (\cref{sec:ablation}) for a more detailed analysis.

\section{Experiments}
\label{sec:exp}

\subsection{Setup}
We compare \alg with two state-of-the-art methods:
COLMAP~\cite{schoenberger2016sfm} 
(commit \href{https://github.com/colmap/colmap/commit/c4a3b308bddf391b4e7c62e835720f83c13aea8b}{\small \texttt{c4a3b30}})
and GLOMAP~\cite{pan2024glomap} 
(commit \href{https://github.com/colmap/glomap/commit/01060b4be509902ea9aaefaae982a2cb941ae5c3}{\small \texttt{01060b4}}), both with GPU-accelerated Ceres~\cite{Agarwal_Ceres_Solver_2022} solver enabled.
For all three methods, we use the COLMAP image matching system. We 
use shared intrinsics if all images in a scene are from the same camera, and leave all other hyper-parameters at their default values in COLMAP and GLOMAP. We run all three methods on a machine with a single A6000 (Ampere) GPU and an AMD EPYC 9274F CPU (Zen 4) with 24 cores / 48 threads. By default, \alg uses 2 CPU threads, whereas COLMAP and GLOMAP use all 48 threads. We report more detailed speed comparisons with different hardware configurations in \cref{tab:detailed_timing} in the appendix.

\begin{table*}[t]
    \resizebox{1.\linewidth}{!}{
    \tablestyle{2pt}{1.1}
\begin{tabular}{r r  r@{~}r@{~}r | r@{~}r@{~}r | r@{~}r@{~}r | r@{~}r@{~}r | r@{~}r@{~}r | r@{~}r@{~}r}

\toprule

  &  &  \multicolumn{3}{c}{time (sec)} &
\multicolumn{3}{c}{ATE$\downarrow$} &
\multicolumn{3}{c}{RTA@3$\uparrow$} &
\multicolumn{3}{c}{AUC-R\&T @ 3 $\uparrow$} &
\multicolumn{3}{c}{RTA@1$\uparrow$} &
\multicolumn{3}{c}{AUC-R\&T @ 1 $\uparrow$} \\

\cmidrule(lr){3-5} \cmidrule(lr){6-8} \cmidrule(lr){9-11}
\cmidrule(lr){12-14} \cmidrule(lr){15-17} \cmidrule(lr){18-20}

  & n\_imgs &
\multicolumn{1}{c}{\scriptsize \alg} & \multicolumn{1}{c}{\scriptsize GLOMAP} & \multicolumn{1}{c}{\scriptsize COLMAP} &
\multicolumn{1}{c}{\scriptsize \alg} & \multicolumn{1}{c}{\scriptsize GLOMAP} & \multicolumn{1}{c}{\scriptsize COLMAP} &
\multicolumn{1}{c}{\scriptsize \alg} & \multicolumn{1}{c}{\scriptsize GLOMAP} & \multicolumn{1}{c}{\scriptsize COLMAP} &
\multicolumn{1}{c}{\scriptsize \alg} & \multicolumn{1}{c}{\scriptsize GLOMAP} & \multicolumn{1}{c}{\scriptsize COLMAP} &
\multicolumn{1}{c}{\scriptsize \alg} & \multicolumn{1}{c}{\scriptsize GLOMAP} & \multicolumn{1}{c}{\scriptsize COLMAP} &
\multicolumn{1}{c}{\scriptsize \alg} & \multicolumn{1}{c}{\scriptsize GLOMAP} & \multicolumn{1}{c}{\scriptsize COLMAP} \\

\midrule
     mipnerf360  (9) & 215.6 &  \cellCD[24pt]{69a84f}{33} &  \cellCD[24pt]{b7d7a8}{165} &  \cellCD[24pt]{ffffff}{503} &  \cellCD[22pt]{ffffff}{4.2e-4} &  \cellCD[22pt]{69a84f}{3.3e-5} &  \cellCD[22pt]{b7d7a8}{5.8e-5} &  \cellCD[22pt]{b7d7a8}{99.9} &  \cellCD[22pt]{b7d7a8}{100.0} &  \cellCD[22pt]{b7d7a8}{100.0} &  \cellCD[22pt]{b7d7a8}{97.4} &  \cellCD[22pt]{b7d7a8}{98.2} &  \cellCD[22pt]{b7d7a8}{97.2} &  \cellCD[22pt]{b7d7a8}{99.8} &  \cellCD[22pt]{b7d7a8}{100.0} &  \cellCD[22pt]{b7d7a8}{99.7} &  \cellCD[22pt]{b7d7a8}{92.4} &  \cellCD[22pt]{69a84f}{94.6} &  \cellCD[22pt]{b7d7a8}{91.9} \\
  tnt\_advanced  (6) & 337.8 &  \cellCD[24pt]{69a84f}{61} &  \cellCD[24pt]{b7d7a8}{357} &  \cellCD[24pt]{ffffff}{1016} &  \cellCD[22pt]{b7d7a8}{6.4e-3} &  \cellCD[22pt]{ffffff}{1.2e-2} &  \cellCD[22pt]{69a84f}{1.2e-3} &  \cellCD[22pt]{ffffff}{71.4} &  \cellCD[22pt]{b7d7a8}{79.1} &  \cellCD[22pt]{69a84f}{98.5} &  \cellCD[22pt]{ffffff}{42.6} &  \cellCD[22pt]{b7d7a8}{75.3} &  \cellCD[22pt]{69a84f}{94.8} &  \cellCD[22pt]{ffffff}{42.3} &  \cellCD[22pt]{b7d7a8}{77.5} &  \cellCD[22pt]{69a84f}{97.0} &  \cellCD[22pt]{ffffff}{16.7} &  \cellCD[22pt]{b7d7a8}{69.8} &  \cellCD[22pt]{69a84f}{90.0} \\
tnt\_intermediate  (8) & 268.6 &  \cellCD[24pt]{69a84f}{35} &  \cellCD[24pt]{b7d7a8}{314} &  \cellCD[24pt]{ffffff}{833} &  \cellCD[22pt]{b7d7a8}{7.8e-5} &  \cellCD[22pt]{69a84f}{1.9e-5} &  \cellCD[22pt]{ffffff}{2.6e-4} &  \cellCD[22pt]{b7d7a8}{99.9} &  \cellCD[22pt]{b7d7a8}{100.0} &  \cellCD[22pt]{b7d7a8}{99.8} &  \cellCD[22pt]{ffffff}{94.1} &  \cellCD[22pt]{b7d7a8}{99.0} &  \cellCD[22pt]{b7d7a8}{98.9} &  \cellCD[22pt]{b7d7a8}{99.3} &  \cellCD[22pt]{b7d7a8}{99.9} &  \cellCD[22pt]{b7d7a8}{99.5} &  \cellCD[22pt]{ffffff}{83.1} &  \cellCD[22pt]{b7d7a8}{96.9} &  \cellCD[22pt]{b7d7a8}{97.3} \\
  tnt\_training  (7) & 470.1 &  \cellCD[24pt]{69a84f}{63} &  \cellCD[24pt]{b7d7a8}{515} &  \cellCD[24pt]{ffffff}{2751} &  \cellCD[22pt]{b7d7a8}{3.0e-3} &  \cellCD[22pt]{ffffff}{1.1e-2} &  \cellCD[22pt]{69a84f}{3.0e-4} &  \cellCD[22pt]{b7d7a8}{87.8} &  \cellCD[22pt]{b7d7a8}{88.7} &  \cellCD[22pt]{69a84f}{99.9} &  \cellCD[22pt]{ffffff}{77.2} &  \cellCD[22pt]{b7d7a8}{87.9} &  \cellCD[22pt]{69a84f}{99.5} &  \cellCD[22pt]{ffffff}{82.1} &  \cellCD[22pt]{b7d7a8}{88.6} &  \cellCD[22pt]{69a84f}{99.9} &  \cellCD[22pt]{ffffff}{60.5} &  \cellCD[22pt]{b7d7a8}{86.3} &  \cellCD[22pt]{69a84f}{98.7} \\
      nerf\_osr  (8) & 402.8 &  \cellCD[24pt]{69a84f}{50} &  \cellCD[24pt]{b7d7a8}{324} &  \cellCD[24pt]{ffffff}{3163} &  \cellCD[22pt]{b7d7a8}{1.6e-3} &  \cellCD[22pt]{b7d7a8}{1.1e-3} &  \cellCD[22pt]{b7d7a8}{1.3e-3} &  \cellCD[22pt]{b7d7a8}{91.7} &  \cellCD[22pt]{b7d7a8}{92.0} &  \cellCD[22pt]{b7d7a8}{92.1} &  \cellCD[22pt]{b7d7a8}{70.9} &  \cellCD[22pt]{b7d7a8}{71.9} &  \cellCD[22pt]{b7d7a8}{71.7} &  \cellCD[22pt]{b7d7a8}{71.1} &  \cellCD[22pt]{b7d7a8}{71.9} &  \cellCD[22pt]{b7d7a8}{71.7} &  \cellCD[22pt]{b7d7a8}{43.2} &  \cellCD[22pt]{b7d7a8}{45.2} &  \cellCD[22pt]{b7d7a8}{44.7} \\
  drone\_deploy  (9) & 524.7 &  \cellCD[24pt]{69a84f}{91} &  \cellCD[24pt]{b7d7a8}{365} &  \cellCD[24pt]{ffffff}{3352} &  \cellCD[22pt]{b7d7a8}{4.9e-3} &  \cellCD[22pt]{b7d7a8}{4.3e-3} &  \cellCD[22pt]{69a84f}{2.0e-3} &  \cellCD[22pt]{b7d7a8}{97.9} &  \cellCD[22pt]{b7d7a8}{98.2} &  \cellCD[22pt]{ffffff}{91.3} &  \cellCD[22pt]{b7d7a8}{79.2} &  \cellCD[22pt]{b7d7a8}{81.1} &  \cellCD[22pt]{ffffff}{65.2} &  \cellCD[22pt]{b7d7a8}{89.6} &  \cellCD[22pt]{b7d7a8}{91.5} &  \cellCD[22pt]{ffffff}{73.5} &  \cellCD[22pt]{b7d7a8}{50.4} &  \cellCD[22pt]{69a84f}{53.5} &  \cellCD[22pt]{ffffff}{40.2} \\
        zipnerf  (4) & 1527.2 &  \cellCD[24pt]{69a84f}{119} &  \cellCD[24pt]{b7d7a8}{690} &  \cellCD[24pt]{ffffff}{3820} &  \cellCD[22pt]{b7d7a8}{3.0e-3} &  \cellCD[22pt]{ffffff}{7.1e-3} &  \cellCD[22pt]{69a84f}{3.4e-4} &  \cellCD[22pt]{b7d7a8}{99.0} &  \cellCD[22pt]{b7d7a8}{98.1} &  \cellCD[22pt]{b7d7a8}{99.7} &  \cellCD[22pt]{ffffff}{92.6} &  \cellCD[22pt]{b7d7a8}{96.6} &  \cellCD[22pt]{b7d7a8}{98.1} &  \cellCD[22pt]{b7d7a8}{97.4} &  \cellCD[22pt]{b7d7a8}{98.0} &  \cellCD[22pt]{b7d7a8}{99.6} &  \cellCD[22pt]{ffffff}{81.4} &  \cellCD[22pt]{b7d7a8}{93.6} &  \cellCD[22pt]{b7d7a8}{95.2} \\
   urban\_scene  (3) & 3824 &  \cellCD[24pt]{69a84f}{515} &  \cellCD[24pt]{b7d7a8}{3664} &  \cellCD[24pt]{ffffff}{61622} &  \cellCD[22pt]{b7d7a8}{1.7e-5} &  \cellCD[22pt]{b7d7a8}{1.4e-5} &  \cellCD[22pt]{b7d7a8}{1.4e-5} &  \cellCD[22pt]{b7d7a8}{99.9} &  \cellCD[22pt]{b7d7a8}{99.9} &  \cellCD[22pt]{b7d7a8}{100.0} &  \cellCD[22pt]{b7d7a8}{95.3} &  \cellCD[22pt]{b7d7a8}{97.0} &  \cellCD[22pt]{b7d7a8}{97.0} &  \cellCD[22pt]{b7d7a8}{99.5} &  \cellCD[22pt]{b7d7a8}{99.6} &  \cellCD[22pt]{b7d7a8}{99.6} &  \cellCD[22pt]{ffffff}{86.3} &  \cellCD[22pt]{b7d7a8}{91.2} &  \cellCD[22pt]{b7d7a8}{91.3} \\
mill19\_building  & 1920 &  \cellCD[24pt]{69a84f}{258} &  \cellCD[24pt]{b7d7a8}{6289} &  \cellCD[24pt]{ffffff}{27080} &  \cellCD[22pt]{b7d7a8}{3.0e-4} &  \cellCD[22pt]{ffffff}{1.3e-2} &  \cellCD[22pt]{69a84f}{1.9e-5} &  \cellCD[22pt]{b7d7a8}{99.9} &  \cellCD[22pt]{f4cccc}{0.1} &  \cellCD[22pt]{b7d7a8}{99.9} &  \cellCD[22pt]{b7d7a8}{95.5} &  \cellCD[22pt]{f4cccc}{0.0} &  \cellCD[22pt]{b7d7a8}{95.6} &  \cellCD[22pt]{b7d7a8}{99.3} &  \cellCD[22pt]{f4cccc}{0.0} &  \cellCD[22pt]{b7d7a8}{99.3} &  \cellCD[22pt]{b7d7a8}{87.0} &  \cellCD[22pt]{f4cccc}{0.0} &  \cellCD[22pt]{b7d7a8}{87.4} \\
 mill19\_rubble  & 1657 &  \cellCD[24pt]{69a84f}{240} &  \cellCD[24pt]{b7d7a8}{2849} &  \cellCD[24pt]{ffffff}{12153} &  \cellCD[22pt]{b7d7a8}{3.6e-5} &  \cellCD[22pt]{ffffff}{6.4e-5} &  \cellCD[22pt]{b7d7a8}{3.4e-5} &  \cellCD[22pt]{b7d7a8}{99.9} &  \cellCD[22pt]{b7d7a8}{99.8} &  \cellCD[22pt]{b7d7a8}{99.9} &  \cellCD[22pt]{b7d7a8}{93.6} &  \cellCD[22pt]{b7d7a8}{94.5} &  \cellCD[22pt]{b7d7a8}{94.6} &  \cellCD[22pt]{b7d7a8}{98.6} &  \cellCD[22pt]{b7d7a8}{98.6} &  \cellCD[22pt]{b7d7a8}{98.7} &  \cellCD[22pt]{ffffff}{81.6} &  \cellCD[22pt]{b7d7a8}{84.7} &  \cellCD[22pt]{b7d7a8}{84.8} \\
eyeful\_apartment  & 3804 &  \cellCD[24pt]{69a84f}{549} &  \cellCD[24pt]{b7d7a8}{5905} &  \cellCD[24pt]{ffffff}{185361} &  \cellCD[22pt]{b7d7a8}{2.8e-3} &  \cellCD[22pt]{ffffff}{9.4e-3} &  \cellCD[22pt]{b7d7a8}{2.2e-3} &  \cellCD[22pt]{b7d7a8}{86.8} &  \cellCD[22pt]{ffffff}{75.0} &  \cellCD[22pt]{69a84f}{90.2} &  \cellCD[22pt]{ffffff}{45.5} &  \cellCD[22pt]{b7d7a8}{50.5} &  \cellCD[22pt]{69a84f}{62.0} &  \cellCD[22pt]{ffffff}{51.1} &  \cellCD[22pt]{b7d7a8}{61.3} &  \cellCD[22pt]{69a84f}{71.7} &  \cellCD[22pt]{ffffff}{6.4} &  \cellCD[22pt]{b7d7a8}{18.2} &  \cellCD[22pt]{69a84f}{21.9} \\
eyeful\_kitchen  & 6042 &  \cellCD[24pt]{69a84f}{2202} &  \cellCD[24pt]{b7d7a8}{22884} &  \cellCD[24pt]{c0c0c0}{timeout} &  \cellCD[22pt]{69a84f}{3.1e-3} &  \cellCD[22pt]{b7d7a8}{7.4e-3} &  \cellCD[22pt]{c0c0c0}{\vphantom{0}-} &  \cellCD[22pt]{69a84f}{85.0} &  \cellCD[22pt]{b7d7a8}{59.9} &  \cellCD[22pt]{c0c0c0}{\vphantom{0}-} &  \cellCD[22pt]{b7d7a8}{38.1} &  \cellCD[22pt]{69a84f}{41.2} &  \cellCD[22pt]{c0c0c0}{\vphantom{0}-} &  \cellCD[22pt]{b7d7a8}{46.7} &  \cellCD[22pt]{69a84f}{51.7} &  \cellCD[22pt]{c0c0c0}{\vphantom{0}-} &  \cellCD[22pt]{b7d7a8}{4.6} &  \cellCD[22pt]{69a84f}{14.4} &  \cellCD[22pt]{c0c0c0}{\vphantom{0}-} \\
\bottomrule
\end{tabular}

    }
    \caption{Speed and pose accuracy of \alg, GLOMAP, and COLMAP. All three methods are accelerated by GPU. For datasets with more than two scenes, we denote the average metrics as \texttt{dataset-name(\#scenes)}. In particular, Tanks and Temples~\cite{knapitsch2017tanks} has three official splits, and we do the averaging separately for them. Mill-19~\cite{turki2022mega} and Eyeful Tower~\cite{VRNeRF} scenes are listed separately. Metrics are color-coded in {\color{teal}green}, with color changes if the percentage gap $>$$2\%$ or ATE ratio $>$$1.5$. {\color{red}Red} denotes complete failures and {\color{gray}gray} means the method did not finish in a week. Note the significant speedup of \alg vs. previous work, especially on larger scenes.}
    \label{tab:pose_overall}

    \vspace{-5mm}
\end{table*}
\begin{table}[t]
\centering
\resizebox{1.0\linewidth}{!}{
    \tablestyle{4pt}{1.1}
\begin{tabular}{l l | ccc | cc}
\toprule
  &  & \multicolumn{3}{c}{Absolute PSNR $\uparrow$} & \multicolumn{2}{c}{Relative to COLMAP} \\
\cmidrule(lr){3-5}  \cmidrule(lr){6-7}
  &  & \alg & GLOMAP & COLMAP & \alg & GLOMAP \\
\midrule
m360\_bicycle & { Zip-NeRF} &  \cellCD[18pt]{b7d7a8}{25.60} &  \cellCD[18pt]{b7d7a8}{25.78} &  \cellCD[18pt]{b7d7a8}{25.86} &  \cellCD[18pt]{f4cccc}{-0.26} &  \cellCD[18pt]{ffffff}{-0.08} \\
 & { + CamP} &  \cellCD[18pt]{b7d7a8}{26.21} &  \cellCD[18pt]{b7d7a8}{26.36} &  \cellCD[18pt]{b7d7a8}{26.41} &  \cellCD[18pt]{ffffff}{-0.21} &  \cellCD[18pt]{ffffff}{-0.05} \\
\cmidrule(lr){2-7}
 & { GSplat} &  \cellCD[18pt]{b7d7a8}{25.51} &  \cellCD[18pt]{b7d7a8}{25.59} &  \cellCD[18pt]{b7d7a8}{25.62} &  \cellCD[18pt]{ffffff}{-0.11} &  \cellCD[18pt]{ffffff}{-0.03} \\
\midrule
m360\_bonsai & { Zip-NeRF} &  \cellCD[18pt]{b7d7a8}{34.78} &  \cellCD[18pt]{b7d7a8}{34.91} &  \cellCD[18pt]{ffffff}{34.47} &  \cellCD[18pt]{fff2cc}{0.31} &  \cellCD[18pt]{fff2cc}{0.44} \\
 & { + CamP} &  \cellCD[18pt]{b7d7a8}{35.26} &  \cellCD[18pt]{b7d7a8}{35.32} &  \cellCD[18pt]{b7d7a8}{35.37} &  \cellCD[18pt]{ffffff}{-0.12} &  \cellCD[18pt]{ffffff}{-0.05} \\
\cmidrule(lr){2-7}
 & { GSplat} &  \cellCD[18pt]{b7d7a8}{32.32} &  \cellCD[18pt]{b7d7a8}{32.29} &  \cellCD[18pt]{ffffff}{31.49} &  \cellCD[18pt]{fff2cc}{0.84} &  \cellCD[18pt]{fff2cc}{0.81} \\
\midrule
m360\_counter & { Zip-NeRF} &  \cellCD[18pt]{b7d7a8}{28.97} &  \cellCD[18pt]{b7d7a8}{28.95} &  \cellCD[18pt]{b7d7a8}{29.18} &  \cellCD[18pt]{ffffff}{-0.21} &  \cellCD[18pt]{ffffff}{-0.23} \\
 & { + CamP} &  \cellCD[18pt]{b7d7a8}{29.09} &  \cellCD[18pt]{b7d7a8}{29.18} &  \cellCD[18pt]{b7d7a8}{29.29} &  \cellCD[18pt]{ffffff}{-0.20} &  \cellCD[18pt]{ffffff}{-0.12} \\
\cmidrule(lr){2-7}
 & { GSplat} &  \cellCD[18pt]{b7d7a8}{28.99} &  \cellCD[18pt]{b7d7a8}{29.06} &  \cellCD[18pt]{b7d7a8}{29.02} &  \cellCD[18pt]{ffffff}{-0.02} &  \cellCD[18pt]{ffffff}{0.04} \\
\midrule
m360\_flowers & { Zip-NeRF} &  \cellCD[18pt]{b7d7a8}{22.05} &  \cellCD[18pt]{b7d7a8}{22.29} &  \cellCD[18pt]{b7d7a8}{21.89} &  \cellCD[18pt]{ffffff}{0.15} &  \cellCD[18pt]{fff2cc}{0.40} \\
 & { + CamP} &  \cellCD[18pt]{b7d7a8}{23.53} &  \cellCD[18pt]{b7d7a8}{23.47} &  \cellCD[18pt]{b7d7a8}{23.27} &  \cellCD[18pt]{fff2cc}{0.25} &  \cellCD[18pt]{ffffff}{0.20} \\
\cmidrule(lr){2-7}
 & { GSplat} &  \cellCD[18pt]{b7d7a8}{21.74} &  \cellCD[18pt]{b7d7a8}{21.79} &  \cellCD[18pt]{b7d7a8}{21.59} &  \cellCD[18pt]{ffffff}{0.15} &  \cellCD[18pt]{ffffff}{0.20} \\
\midrule
m360\_garden & { Zip-NeRF} &  \cellCD[18pt]{b7d7a8}{28.10} &  \cellCD[18pt]{b7d7a8}{28.20} &  \cellCD[18pt]{b7d7a8}{28.20} &  \cellCD[18pt]{ffffff}{-0.11} &  \cellCD[18pt]{ffffff}{0.00} \\
 & { + CamP} &  \cellCD[18pt]{b7d7a8}{28.54} &  \cellCD[18pt]{b7d7a8}{28.49} &  \cellCD[18pt]{b7d7a8}{28.54} &  \cellCD[18pt]{ffffff}{0.00} &  \cellCD[18pt]{ffffff}{-0.05} \\
\cmidrule(lr){2-7}
 & { GSplat} &  \cellCD[18pt]{b7d7a8}{27.61} &  \cellCD[18pt]{b7d7a8}{27.67} &  \cellCD[18pt]{b7d7a8}{27.72} &  \cellCD[18pt]{ffffff}{-0.11} &  \cellCD[18pt]{ffffff}{-0.05} \\
\midrule
m360\_kitchen & { Zip-NeRF} &  \cellCD[18pt]{b7d7a8}{32.29} &  \cellCD[18pt]{b7d7a8}{32.43} &  \cellCD[18pt]{b7d7a8}{32.31} &  \cellCD[18pt]{ffffff}{-0.02} &  \cellCD[18pt]{ffffff}{0.12} \\
 & { + CamP} &  \cellCD[18pt]{69a84f}{32.47} &  \cellCD[18pt]{b7d7a8}{32.19} &  \cellCD[18pt]{b7d7a8}{32.21} &  \cellCD[18pt]{fff2cc}{0.27} &  \cellCD[18pt]{ffffff}{-0.02} \\
\cmidrule(lr){2-7}
 & { GSplat} &  \cellCD[18pt]{b7d7a8}{31.36} &  \cellCD[18pt]{b7d7a8}{31.62} &  \cellCD[18pt]{b7d7a8}{31.58} &  \cellCD[18pt]{ffffff}{-0.22} &  \cellCD[18pt]{ffffff}{0.05} \\
\midrule
m360\_room & { Zip-NeRF} &  \cellCD[18pt]{b7d7a8}{32.81} &  \cellCD[18pt]{b7d7a8}{32.94} &  \cellCD[18pt]{b7d7a8}{32.93} &  \cellCD[18pt]{ffffff}{-0.12} &  \cellCD[18pt]{ffffff}{0.01} \\
 & { + CamP} &  \cellCD[18pt]{b7d7a8}{32.51} &  \cellCD[18pt]{b7d7a8}{32.48} &  \cellCD[18pt]{b7d7a8}{32.44} &  \cellCD[18pt]{ffffff}{0.07} &  \cellCD[18pt]{ffffff}{0.04} \\
\cmidrule(lr){2-7}
 & { GSplat} &  \cellCD[18pt]{b7d7a8}{31.77} &  \cellCD[18pt]{b7d7a8}{31.71} &  \cellCD[18pt]{b7d7a8}{31.67} &  \cellCD[18pt]{ffffff}{0.11} &  \cellCD[18pt]{ffffff}{0.04} \\
\midrule
m360\_stump & { Zip-NeRF} &  \cellCD[18pt]{b7d7a8}{27.34} &  \cellCD[18pt]{b7d7a8}{27.41} &  \cellCD[18pt]{b7d7a8}{27.43} &  \cellCD[18pt]{ffffff}{-0.09} &  \cellCD[18pt]{ffffff}{-0.02} \\
 & { + CamP} &  \cellCD[18pt]{b7d7a8}{28.10} &  \cellCD[18pt]{b7d7a8}{28.03} &  \cellCD[18pt]{b7d7a8}{28.03} &  \cellCD[18pt]{ffffff}{0.07} &  \cellCD[18pt]{ffffff}{0.00} \\
\cmidrule(lr){2-7}
 & { GSplat} &  \cellCD[18pt]{b7d7a8}{26.97} &  \cellCD[18pt]{b7d7a8}{26.89} &  \cellCD[18pt]{b7d7a8}{26.84} &  \cellCD[18pt]{ffffff}{0.13} &  \cellCD[18pt]{ffffff}{0.05} \\
\midrule
m360\_treehill & { Zip-NeRF} &  \cellCD[18pt]{ffffff}{23.73} &  \cellCD[18pt]{b7d7a8}{24.05} &  \cellCD[18pt]{b7d7a8}{24.04} &  \cellCD[18pt]{f4cccc}{-0.31} &  \cellCD[18pt]{ffffff}{0.01} \\
 & { + CamP} &  \cellCD[18pt]{b7d7a8}{25.73} &  \cellCD[18pt]{b7d7a8}{25.74} &  \cellCD[18pt]{b7d7a8}{25.99} &  \cellCD[18pt]{f4cccc}{-0.26} &  \cellCD[18pt]{ffffff}{-0.25} \\
\cmidrule(lr){2-7}
 & { GSplat} &  \cellCD[18pt]{b7d7a8}{22.71} &  \cellCD[18pt]{b7d7a8}{22.88} &  \cellCD[18pt]{b7d7a8}{22.80} &  \cellCD[18pt]{ffffff}{-0.08} &  \cellCD[18pt]{ffffff}{0.08} \\
\midrule
tnt\_training (7) & { InstantNGP} &  \cellCD[18pt]{b7d7a8}{20.73} &  \cellCD[18pt]{ffffff}{19.37} &  \cellCD[18pt]{69a84f}{21.05} &  \cellCD[18pt]{f4cccc}{-0.32} &  \cellCD[18pt]{f4cccc}{-1.68} \\
\cmidrule(lr){2-7}
 & { GSplat} &  \cellCD[18pt]{b7d7a8}{23.22} &  \cellCD[18pt]{ffffff}{21.54} &  \cellCD[18pt]{69a84f}{24.19} &  \cellCD[18pt]{f4cccc}{-0.97} &  \cellCD[18pt]{f4cccc}{-2.65} \\
\midrule
tnt\_intermediate (8) & { InstantNGP} &  \cellCD[18pt]{b7d7a8}{22.29} &  \cellCD[18pt]{b7d7a8}{22.51} &  \cellCD[18pt]{b7d7a8}{22.38} &  \cellCD[18pt]{ffffff}{-0.09} &  \cellCD[18pt]{ffffff}{0.13} \\
\cmidrule(lr){2-7}
 & { GSplat} &  \cellCD[18pt]{ffffff}{24.24} &  \cellCD[18pt]{b7d7a8}{25.28} &  \cellCD[18pt]{b7d7a8}{25.24} &  \cellCD[18pt]{f4cccc}{-1.00} &  \cellCD[18pt]{ffffff}{0.03} \\
\midrule
tnt\_advanced (6) & { InstantNGP} &  \cellCD[18pt]{ffffff}{16.59} &  \cellCD[18pt]{b7d7a8}{16.94} &  \cellCD[18pt]{69a84f}{17.55} &  \cellCD[18pt]{f4cccc}{-0.95} &  \cellCD[18pt]{f4cccc}{-0.60} \\
\cmidrule(lr){2-7}
 & { GSplat} &  \cellCD[18pt]{b7d7a8}{18.82} &  \cellCD[18pt]{b7d7a8}{18.74} &  \cellCD[18pt]{69a84f}{22.06} &  \cellCD[18pt]{f4cccc}{-3.24} &  \cellCD[18pt]{f4cccc}{-3.32} \\
\bottomrule
\end{tabular}

}
\caption{
Novel view synthesis evaluation on MipNeRF360~\cite{barron2022mipnerf360} and Tanks and Temples~\cite{knapitsch2017tanks}.  Results for MipNeRF360 are listed separately for each scene, and those for Tanks and Temples are averaged over all scenes in each of the three splits. The color changes only if the PSNR difference $>$$0.25$. We report results for Zip-NeRF, Zip-NeRF + CamP optimization, and Gaussian Splatting. 
}
\label{tab:nerf_m360}
\end{table}

\begin{table}[t]
\centering
\resizebox{1.0\linewidth}{!}{
\tablestyle{1pt}{1.1}
\begin{tabular}{l ccc@{\,} | ccc@{\,} | ccc}
\toprule
& \multicolumn{3}{c}{tnt\_training (7)} & \multicolumn{3}{c}{tnt\_intermediate (8)} & \multicolumn{3}{c}{tnt\_advanced (6)} \\
\cmidrule(lr){2-4} \cmidrule(lr){5-7} \cmidrule(lr){8-10}
& \scriptsize ATE & \scriptsize RTA@5 & \scriptsize RRA@5
& \scriptsize ATE & \scriptsize RTA@5 & \scriptsize RRA@5
& \scriptsize ATE & \scriptsize RTA@5 & \scriptsize RRA@5 \\
\midrule
ACE-Zero~\cite{brachmann2024scene}      &  \cellCD[22pt]{ffffff}{1.2e-2} &  \cellCD[18pt]{ffffff}{72.9} &  \cellCD[18pt]{ffffff}{73.9} &  \cellCD[22pt]{ffffff}{8.0e-3} &  \cellCD[18pt]{ffffff}{74.0} &  \cellCD[18pt]{ffffff}{67.5} &  \cellCD[22pt]{ffffff}{2.8e-2} &  \cellCD[18pt]{ffffff}{19.1} &  \cellCD[18pt]{ffffff}{22.9}  \\
MAST3R-SfM~\cite{duisterhof2024mast3r}  &  \cellCD[22pt]{b7d7a8}{6.2e-3} &  \cellCD[18pt]{ffffff}{64.9} &  \cellCD[18pt]{ffffff}{56.2} &  \cellCD[22pt]{ffffff}{7.2e-3} &  \cellCD[18pt]{ffffff}{57.5} &  \cellCD[18pt]{ffffff}{50.8} &  \cellCD[22pt]{ffffff}{2.0e-2} &  \cellCD[18pt]{ffffff}{36.5} &  \cellCD[18pt]{ffffff}{38.8}  \\
GLOMAP                                  &  \cellCD[22pt]{ffffff}{1.1e-2} &  \cellCD[18pt]{b7d7a8}{88.8} &  \cellCD[18pt]{b7d7a8}{89.3} &  \cellCD[22pt]{69a84f}{1.9e-5} &  \cellCD[18pt]{b7d7a8}{100.0} &  \cellCD[18pt]{b7d7a8}{100.0} &  \cellCD[22pt]{b7d7a8}{1.2e-2} &  \cellCD[18pt]{69a84f}{79.3} &  \cellCD[18pt]{b7d7a8}{80.5}  \\
\alg                                    &  \cellCD[22pt]{69a84f}{3.2e-3} &  \cellCD[18pt]{b7d7a8}{88.8} &  \cellCD[18pt]{69a84f}{95.8} &  \cellCD[22pt]{b7d7a8}{9.2e-5} &  \cellCD[18pt]{b7d7a8}{100.0} &  \cellCD[18pt]{b7d7a8}{100.0} &  \cellCD[22pt]{69a84f}{6.8e-3} &  \cellCD[18pt]{b7d7a8}{70.5} &  \cellCD[18pt]{b7d7a8}{82.1}  \\
\bottomrule
\end{tabular}

}
\caption{
Comparison to learning-based SfM on Tanks and Temples~\cite{knapitsch2017tanks}. We use COLMAP poses from \citet{kulhanek2024nerfbaselines} as reference. We average numbers for scenes in each split. 
}
\label{tab:compare_learning}
\end{table}

\minihead{Datasets}
We focus on the case of high-overlap images densely connected by feature matching, and evaluate the three methods on eight datasets: 
MipNeRF360~\cite{barron2022mipnerf360}, Tanks and Temples~\cite{knapitsch2017tanks}, ZipNeRF~\cite{barron2023zipnerf}, NeRF-OSR~\cite{rudnev2022nerf}, DroneDeploy~\cite{Pilkington2022}, Mill-19~\cite{turki2022mega}, Urbanscene3D~\cite{lin2022capturing}, and Eyeful Tower~\cite{VRNeRF}. They cover a wide range of real-world scenarios and camera trajectory patterns. The number of images per scene ranges from around 200 to 6000. \cref{app:data} provides a more detailed discussion of the GT poses provided along with these datasets.

\minihead{Metrics}
We report wall-clock time in seconds, excluding the time required for feature extraction and matching (identical for all three methods and dominated by the SfM backend time for large scenes). We evaluate pose accuracy using the standard metrics~\cite{schoenberger2016sfm,pan2024glomap, duisterhof2024mast3r}: ATE, RRA@$\delta$, RTA@$\delta$, and AUC@$\delta$. For some of the scenes, we also evaluate the novel view synthesis quality of NeRF~\cite{mildenhall2021nerf} and Gaussian Splatting~\cite{kerbl20233d} trained on the output poses, intrinsics, and triangulated point clouds.

\subsection{Analysis}

\noindent\textbf{Pose accuracy} Table~\ref{tab:pose_overall} compares the three methods in average camera pose metrics on all the datasets (we include per-scene evaluation results in the supplementary material). In general, our method is much faster than both GLOMAP and COLMAP. The speedup over GLOMAP is less dramatic when there are only a few hundred images (e.g., MipNeRF360), but it can be about $10\times$ faster on scenes with several thousand images (e.g., Urbanscene3D, Mill-19, and Eyeful Tower). On most datasets, \alg is on par with GLOMAP and COLMAP in terms of RTA@$3$. There is a more prominent difference for stricter metrics (RTA@$1$, AUC@$3$, AUC@$1$). This shows that while \alg succeeds in recovering the overall structures of camera trajectories, it does achieve the highest level of precision when the error is reduced to one or two degrees.

None of the methods are perfect. \alg performs particularly bad on the Advanced split of Tanks and Temples, probably because there are many erroneous matches due to repetitive patterns and symmetric structures in the scenes. This is a well-known problem of global SfM (i.e., GLOMAP also suffers a significant drop in performance compared to COLMAP), and incremental SfM methods like COLMAP are more robust in these settings. On the building scene of the Mill-19 dataset, GLOMAP fail catastrophically, however \alg and COLMAP remain highly accurate. On DroneDeploy, none of the three methods is very good in terms of AUC@$1$ and RTA@$1$.

In \cref{tab:compare_learning}, we compare \alg to two representative learning-based methods, ACE-Zero~\cite{brachmann2024scene} and MAST3R-SfM~\cite{duisterhof2024mast3r}, where we include the results from \citet[Tab.~10]{duisterhof2024mast3r}. 
Both methods perform significantly worse than \alg and GLOMAP. This indicates that while learning-based methods are promising, they still lag far behind traditional methods in terms of pose accuracy.

\noindent\textbf{Novel view synthesis} Table~\ref{tab:nerf_m360} evaluates the quality of novel view synthesis on MipNeRF360 and Tanks and Temples when using \alg, COLMAP, and GLOMAP to estimate the camera poses. We use ZipNeRF~\cite{barron2023zipnerf}, a very high-quality NeRF method, for MipNeRF360, and use Instant-NGP~\cite{muller2022instant} for Tank and Temples, which offers a better trade-off between quality and speed. We also evaluate the performance of Gaussian Splatting~\cite{kerbl20233d} on both datasets.

While \alg lags behind GLOMAP and COLMAP on most MipNeRF360 scenes, the PSNR difference is within $0.5$. 
On Tanks and Temples, \alg performs on par with GLOMAP, but both are worse than COLMAP. Here, again, the lower pose accuracy of \alg under the strictest metrics does not prevent \alg poses from yielding competitive PSNR. %
These results suggest that pose accuracy under a strict metric could be a misleading proxy for downstream view synthesis quality, and vice versa.

We also investigate the impact of different SfM poses on rendering with CamP~\cite{park2023camp}, which simultaneously optimizes the radiance field and refines the camera poses. We include the results in \cref{tab:nerf_m360} for comparison. 
In general, CamP improves the PSNR for all the three methods, and for some scenes (e.g., flowers, garden, kitchen, etc.) the gap in rendering quality is closed and sometimes even reversed.

\subsection{Ablations}
\label{sec:ablation}

\minihead{Kernel fusion} In \cref{tab:kernel_fusion}, we show the timing comparison of naive PyTorch and kernel fusion approaches to implementing the first-order optimization of epipolar adjustment. Profiling and comparing the CPU and GPU times for these two approaches is challenging due to the various forms of execution overlap. Instead, we directly compare the wall-clock time on different hardware setups. On small-to-medium scale scenes (i.e., 5k and 50k image pairs), the running time is severely bottlenecked by CPU overhead, and using a slower CPU can significantly impact the speed. Interestingly, the PyTorch version is faster on the less-powerful 2080 Ti than A6000, reflecting that its kernel implementation cannot fully utilize the power of high-end GPUs and is not suitable in our case. Across all the three hardware settings and scene sizes, our fused kernel implementation is around $20\times$ to $90\times$ faster than the naive PyTorch version.
\begin{table}[t]
    \centering
    \resizebox{1.0\linewidth}{!}{
    \tablestyle{8pt}{1.1}
\renewcommand{\arraystretch}{1.0}
\begin{tabular}{l l l rrr}
\toprule
 $\#$ pairs & CPU & GPU &
torch (ms) &
fused (ms) &
speedup \\
\midrule
\multirow{3}{4em}{5k} 
 & 4.05GHz & A6000 & 2.83 & \textbf{0.05} & $56\times$ \\
 & 2.2GHz & A6000 & 9.82 & \textbf{0.11} & $89\times$ \\
 & 2.2GHz & 2080 Ti & 9.41 & \textbf{0.11} & $85\times$ \\
\midrule
\multirow{3}{4em}{50k} 
 & 4.05GHz & A6000 & 8.20 & \textbf{0.14} & $58\times$ \\
 & 2.2GHz & A6000 & 12.47 & \textbf{0.20} & $62\times$ \\
 & 2.2GHz & 2080 Ti & 11.93 & \textbf{0.27} & $44\times$ \\
\midrule
\multirow{3}{4em}{500k} 
 & 4.05GHz & A6000 & 65.94 & \textbf{1.16} & $56\times$ \\
 & 2.2GHz & A6000 & 69.31 & \textbf{1.21} & $53\times$ \\
 & 2.2GHz & 2080 Ti & 44.32 & \textbf{1.92} & $23\times$ \\
\bottomrule
\vspace{-5mm}
\end{tabular}

    }
    \caption{Effect of kernel fusion for epipolar adjustment under different hardware settings and scene sizes ($\#$pairs refers to the number of image pairs). Interestingly, the naive PyTorch implementation is faster on 2080 Ti than A6000 with 500k image pairs, showing that the native PyTorch kernel implementation cannot fully utilize the GPU for our problems. Note that the performance of the same CPU or GPU can be slightly different on different machines.}
    \label{tab:kernel_fusion}
\end{table}

\minihead{Distortion estimation} is one of the first steps of \alg, and its accuracy is critical to the final performance. 
\begin{table}[t!]
    \centering
    \resizebox{1.0\linewidth}{!}{
        \tablestyle{5pt}{1.1}
\begin{tabular}{l  rr @{~~~~} | rr @{~~~~} | rr @{~~~~} | rr}
\toprule
 &
\multicolumn{2}{c}{\scriptsize AUC@3} &
\multicolumn{2}{c}{\scriptsize AUC@10} &
\multicolumn{2}{c}{\scriptsize RTA@3} &
\multicolumn{2}{c}{\scriptsize RTA@10} \\
\cmidrule(lr){2-3} \cmidrule(lr){4-5} \cmidrule(lr){6-7} \cmidrule(lr){8-9}

 & \multicolumn{1}{c}{w/} & \multicolumn{1}{c}{w/o}
& \multicolumn{1}{c}{w/} & \multicolumn{1}{c}{w/o}
& \multicolumn{1}{c}{w/} & \multicolumn{1}{c}{w/o}
& \multicolumn{1}{c}{w/} & \multicolumn{1}{c}{w/o} \\
\midrule
Family & \textbf{95.1} & 72.8 & \textbf{98.5} & 91.8 & \textbf{100.0} & 99.9 & \textbf{100.0} & 100.0 \\
Francis & \textbf{95.5} & 71.1 & \textbf{98.6} & 91.2 & \textbf{99.9} & 99.6 & \textbf{100.0} & 99.9 \\
Horse & \textbf{96.8} & 76.8 & \textbf{99.0} & 93.0 & \textbf{100.0} & 100.0 & \textbf{100.0} & 100.0 \\
Lighthouse & \textbf{90.7} & \cellcolor[HTML]{f4cccc} 4.6 & \textbf{97.1} & \cellcolor[HTML]{f4cccc} 42.2 & \textbf{99.6} & \cellcolor[HTML]{f4cccc} 46.5 & \textbf{100.0} & 98.5 \\
M60 & \textbf{95.6} & \cellcolor[HTML]{f4cccc} 28.3 & \textbf{98.7} & 72.9 & \textbf{99.9} & 85.9 & \textbf{100.0} & 99.7 \\
Panther & \textbf{93.0} & \cellcolor[HTML]{f4cccc} 12.1 & \textbf{97.9} & 64.2 & \textbf{99.9} & 78.3 & \textbf{100.0} & 99.9 \\
Playground & \textbf{84.6} & \cellcolor[HTML]{f4cccc} 2.2 & \textbf{95.4} & \cellcolor[HTML]{f4cccc} 14.4 & \textbf{100.0} & \cellcolor[HTML]{f4cccc} 15.2 & \textbf{100.0} & \cellcolor[HTML]{f4cccc} 51.9 \\
Train & \textbf{92.4} & \cellcolor[HTML]{f4cccc} 54.2 & \textbf{97.7} & 86.0 & \textbf{99.9} & 99.6 & \textbf{100.0} & 99.9 \\
\bottomrule
\end{tabular}

    }
    \caption{Effect of camera distortion estimation on pose accuracy.}
    \label{tab:ablate_distortion}
\end{table}
Table~\ref{tab:ablate_distortion} presents the performance of \alg with and without distortion estimation on the Intermediate split of Tanks and Temples. Without distortion estimation, results drop, sometimes catastrophically. We provide in \cref{app:distortion} an additional insight into the effect of distortion estimate on the immediately following step of focal length estimation.

\minihead{Others} Due to page limit, we put some other ablation results in \cref{app:ablation}, including those for track completion, multiple initialization, and epipolar adjustment.

\section{Limitations and Conclusions}
\label{sec:conclusions}

We introduce \alg, a new structure from motion method focused on simplicity and speed. Contrary to the common practice in other SfM systems, \alg uses first-order optimization extensively and is much faster than state-of-the-art methods (COLMAP and GLOMAP), while achieving comparable performance on pose accuracy and novel view synthesis quality. These improvements do come with a few drawbacks. For example, \alg  might fail on scenes where there are a lot of degenerate motions, and is more sensitive to incorrect matching induced by repetitive patterns and symmetric structures when compared to GLOMAP (please refer to the appendix for a more detailed discussion of limitations). Nevertheless, we believe it is an important step towards highly efficient camera pose estimation for real-world 3D data acquisition at scale.

{
    \small
    \bibliographystyle{ieeenat_fullname}
    \bibliography{main}
}

\clearpage

\appendix
\begin{table*}[]
\resizebox{1.\linewidth}{!}{
    
\tablestyle{4pt}{1.1}
\begin{tabular}{l r dd | ddd | dd }
\toprule
  &  &
\multicolumn{2}{c}{\scriptsize \alg (\textbf{1}G+\textbf{2}C) } & \multicolumn{3}{c}{\scriptsize GLOMAP} & \multicolumn{2}{c}{\scriptsize COLMAP} \\
\cmidrule(lr){3-4} \cmidrule(lr){5-7} \cmidrule(lr){8-9}
  & n\_imgs &
\multicolumn{1}{c}{\scriptsize w/ cuda} &
\multicolumn{1}{c}{\scriptsize w/o cuda } &
\multicolumn{1}{c}{\scriptsize \textbf{1}G+\textbf{48}C} &
\multicolumn{1}{c}{\scriptsize \textbf{1}G+\textbf{12}C} &
\multicolumn{1}{c}{\scriptsize \textbf{48}C} &
\multicolumn{1}{c}{\scriptsize \textbf{1}G+\textbf{48}C} &
\multicolumn{1}{c}{\scriptsize \textbf{48}C}\\
\midrule
     z\_alameda  & 1734 & 134 : \textcolor{teal}{\times 1.0} & 917 : \textcolor{teal}{\times 6.8} & 848 : \textcolor{teal}{\times 6.3} & 934 : \textcolor{teal}{\times 6.9} & 3805 : \textcolor{teal}{\times 28.2} & 4541 : \textcolor{teal}{\times 33.6} & 24641 : \textcolor{teal}{\times 182.5} \\
      z\_berlin  & 1511 & 152 : \textcolor{teal}{\times 1.0} & 545 : \textcolor{teal}{\times 3.6} & 893 : \textcolor{teal}{\times 5.9} & 1015 : \textcolor{teal}{\times 6.7} & 1802 : \textcolor{teal}{\times 11.8} & 6478 : \textcolor{teal}{\times 42.4} & 24648 : \textcolor{teal}{\times 161.4} \\
      z\_london  & 1874 & 102 : \textcolor{teal}{\times 1.0} & 556 : \textcolor{teal}{\times 5.4} & 566 : \textcolor{teal}{\times 5.5} & 669 : \textcolor{teal}{\times 6.5} & 2092 : \textcolor{teal}{\times 20.3} & 2643 : \textcolor{teal}{\times 25.7} & 19238 : \textcolor{teal}{\times 187.1} \\
         z\_nyc  &  990 & 88 : \textcolor{teal}{\times 1.0} & 338 : \textcolor{teal}{\times 3.8} & 451 : \textcolor{teal}{\times 5.1} & 487 : \textcolor{teal}{\times 5.5} & 921 : \textcolor{teal}{\times 10.5} & 1618 : \textcolor{teal}{\times 18.4} & 1988 : \textcolor{teal}{\times 22.6} \\
mill19\_building  & 1920 & 258 : \textcolor{teal}{\times 1.0} & 1366 : \textcolor{teal}{\times 5.3} & 6289 : \textcolor{teal}{\times 24.3} & 7792 : \textcolor{teal}{\times 30.1} & 38428 : \textcolor{teal}{\times 148.4} & 27080 : \textcolor{teal}{\times 104.6} & 152839 : \textcolor{teal}{\times 590.1} \\
 mill19\_rubble  & 1657 & 240 : \textcolor{teal}{\times 1.0} & 789 : \textcolor{teal}{\times 3.3} & 2849 : \textcolor{teal}{\times 11.8} & 2466 : \textcolor{teal}{\times 10.2} & 11571 : \textcolor{teal}{\times 48.0} & 12153 : \textcolor{teal}{\times 50.4} & 64987 : \textcolor{teal}{\times 269.8} \\
   urbn\_Campus  & 5871 & 740 : \textcolor{teal}{\times 1.0} & 3009 : \textcolor{teal}{\times 4.1} & 3869 : \textcolor{teal}{\times 5.2} & 4175 : \textcolor{teal}{\times 5.6} & 21916 : \textcolor{teal}{\times 29.6} & 106055 : \textcolor{teal}{\times 143.2} & 349490 : \textcolor{teal}{\times 472.0} \\
  urbn\_Sci-Art  & 3019 & 445 : \textcolor{teal}{\times 1.0} & 1760 : \textcolor{teal}{\times 4.0} & 4601 : \textcolor{teal}{\times 10.3} & 5712 : \textcolor{teal}{\times 12.8} & 28824 : \textcolor{teal}{\times 64.7} & 42032 : \textcolor{teal}{\times 94.4} & 286454 : \textcolor{teal}{\times 643.4} \\
 eft\_apartment  & 3804 & 549 : \textcolor{teal}{\times 1.0} & 1003 : \textcolor{teal}{\times 1.8} & 5905 : \textcolor{teal}{\times 10.8} & 8341 : \textcolor{teal}{\times 15.2} & 124310 : \textcolor{teal}{\times 226.3} & 185361 : \textcolor{teal}{\times 337.5} & \multicolumn{1}{c}{timeout} \\
   eft\_kitchen  & 6042 & 2202 : \textcolor{teal}{\times 1.0} & 6796 : \textcolor{teal}{\times 3.1} & 22884 : \textcolor{teal}{\times 10.4} & 34287 : \textcolor{teal}{\times 15.6} & \multicolumn{1}{c}{timeout} & \multicolumn{1}{c}{timeout} & \multicolumn{1}{c}{timeout} \\
\bottomrule
\end{tabular}

}
\caption{
Detailed system runtime comparisons (seconds:\textcolor{teal}{speed\_ratio}) with different GPU (G) and CPU threads (C) configurations. Despite the cuda-accelerated ceres solver for bundle adjustment, a significant part of GLOMAP and COLMAP pipeline workload is still CPU-bound, and having at least 12 threads is necessary for higher speed. \alg performs all data structure marshaling on GPU with non-trivial tensor indexing, and consumes less CPU resource. 
}
\label{tab:detailed_timing}
\end{table*}

\begin{table*}[]
\resizebox{1.\linewidth}{!}{
    \tablestyle{2pt}{1.1}
\begin{tabular}{l r  r@{~}r@{~}r | r@{~}r@{~}r | r@{~}r@{~}r | r@{~}r@{~}r | r@{~}r@{~}r | r@{~}r@{~}r | r@{~}r@{~}r | r@{~}r@{~}r }

\toprule

  &  & \multicolumn{3}{c}{time (sec)} &
\multicolumn{3}{c}{ATE$\downarrow$} &
\multicolumn{3}{c}{RTA@3$\uparrow$} &
\multicolumn{3}{c}{RRA@3$\uparrow$} &
\multicolumn{3}{c}{RTA@1$\uparrow$} &
\multicolumn{3}{c}{RRA@1$\uparrow$} &
\multicolumn{3}{c}{AUC-R\&T @ 3 $\uparrow$} &
\multicolumn{3}{c}{AUC-R\&T @ 1 $\uparrow$} \\

\cmidrule(lr){3-5} \cmidrule(lr){6-8} \cmidrule(lr){9-11}
\cmidrule(lr){12-14} \cmidrule(lr){15-17} \cmidrule(lr){18-20}
\cmidrule(lr){21-23} \cmidrule(lr){24-26}

   & n\_imgs &
\multicolumn{1}{c}{\scriptsize \alg} & \multicolumn{1}{c}{\scriptsize GLOMAP} & \multicolumn{1}{c}{\scriptsize COLMAP} &
\multicolumn{1}{c}{\scriptsize \alg} & \multicolumn{1}{c}{\scriptsize GLOMAP} & \multicolumn{1}{c}{\scriptsize COLMAP} &
\multicolumn{1}{c}{\scriptsize \alg} & \multicolumn{1}{c}{\scriptsize GLOMAP} & \multicolumn{1}{c}{\scriptsize COLMAP} &
\multicolumn{1}{c}{\scriptsize \alg} & \multicolumn{1}{c}{\scriptsize GLOMAP} & \multicolumn{1}{c}{\scriptsize COLMAP} &
\multicolumn{1}{c}{\scriptsize \alg} & \multicolumn{1}{c}{\scriptsize GLOMAP} & \multicolumn{1}{c}{\scriptsize COLMAP} &
\multicolumn{1}{c}{\scriptsize \alg} & \multicolumn{1}{c}{\scriptsize GLOMAP} & \multicolumn{1}{c}{\scriptsize COLMAP} &
\multicolumn{1}{c}{\scriptsize \alg} & \multicolumn{1}{c}{\scriptsize GLOMAP} & \multicolumn{1}{c}{\scriptsize COLMAP} &
\multicolumn{1}{c}{\scriptsize \alg} & \multicolumn{1}{c}{\scriptsize GLOMAP} & \multicolumn{1}{c}{\scriptsize COLMAP} \\
\bottomrule
\end{tabular}

}
\caption{Per scene camera pose metrics on the MipNeRF360 Dataset.}
\label{tab:supp_pose_m360}
\end{table*}

\begin{table*}[]
\resizebox{1.\linewidth}{!}{
    \tablestyle{2pt}{1.1}
\begin{tabular}{l r  r@{~}r@{~}r | r@{~}r@{~}r | r@{~}r@{~}r | r@{~}r@{~}r | r@{~}r@{~}r | r@{~}r@{~}r | r@{~}r@{~}r | r@{~}r@{~}r }

\toprule

  &  & \multicolumn{3}{c}{time (sec)} &
\multicolumn{3}{c}{ATE$\downarrow$} &
\multicolumn{3}{c}{RTA@3$\uparrow$} &
\multicolumn{3}{c}{RRA@3$\uparrow$} &
\multicolumn{3}{c}{RTA@1$\uparrow$} &
\multicolumn{3}{c}{RRA@1$\uparrow$} &
\multicolumn{3}{c}{AUC-R\&T @ 3 $\uparrow$} &
\multicolumn{3}{c}{AUC-R\&T @ 1 $\uparrow$} \\

\cmidrule(lr){3-5} \cmidrule(lr){6-8} \cmidrule(lr){9-11}
\cmidrule(lr){12-14} \cmidrule(lr){15-17} \cmidrule(lr){18-20}
\cmidrule(lr){21-23} \cmidrule(lr){24-26}

   & n\_imgs &
\multicolumn{1}{c}{\scriptsize \alg} & \multicolumn{1}{c}{\scriptsize GLOMAP} & \multicolumn{1}{c}{\scriptsize COLMAP} &
\multicolumn{1}{c}{\scriptsize \alg} & \multicolumn{1}{c}{\scriptsize GLOMAP} & \multicolumn{1}{c}{\scriptsize COLMAP} &
\multicolumn{1}{c}{\scriptsize \alg} & \multicolumn{1}{c}{\scriptsize GLOMAP} & \multicolumn{1}{c}{\scriptsize COLMAP} &
\multicolumn{1}{c}{\scriptsize \alg} & \multicolumn{1}{c}{\scriptsize GLOMAP} & \multicolumn{1}{c}{\scriptsize COLMAP} &
\multicolumn{1}{c}{\scriptsize \alg} & \multicolumn{1}{c}{\scriptsize GLOMAP} & \multicolumn{1}{c}{\scriptsize COLMAP} &
\multicolumn{1}{c}{\scriptsize \alg} & \multicolumn{1}{c}{\scriptsize GLOMAP} & \multicolumn{1}{c}{\scriptsize COLMAP} &
\multicolumn{1}{c}{\scriptsize \alg} & \multicolumn{1}{c}{\scriptsize GLOMAP} & \multicolumn{1}{c}{\scriptsize COLMAP} &
\multicolumn{1}{c}{\scriptsize \alg} & \multicolumn{1}{c}{\scriptsize GLOMAP} & \multicolumn{1}{c}{\scriptsize COLMAP} \\
\bottomrule
\end{tabular}

}
\caption{Per scene camera pose metrics on the Tanks and Temples Dataset.}
\label{tab:supp_pose_tnt}
\end{table*}

\begin{table*}[]
\resizebox{1.\linewidth}{!}{
    \tablestyle{2pt}{1.1}
\begin{tabular}{l r  r@{~}r@{~}r | r@{~}r@{~}r | r@{~}r@{~}r | r@{~}r@{~}r | r@{~}r@{~}r | r@{~}r@{~}r | r@{~}r@{~}r | r@{~}r@{~}r }

\toprule

  &  & \multicolumn{3}{c}{time (sec)} &
\multicolumn{3}{c}{ATE$\downarrow$} &
\multicolumn{3}{c}{RTA@3$\uparrow$} &
\multicolumn{3}{c}{RRA@3$\uparrow$} &
\multicolumn{3}{c}{RTA@1$\uparrow$} &
\multicolumn{3}{c}{RRA@1$\uparrow$} &
\multicolumn{3}{c}{AUC-R\&T @ 3 $\uparrow$} &
\multicolumn{3}{c}{AUC-R\&T @ 1 $\uparrow$} \\

\cmidrule(lr){3-5} \cmidrule(lr){6-8} \cmidrule(lr){9-11}
\cmidrule(lr){12-14} \cmidrule(lr){15-17} \cmidrule(lr){18-20}
\cmidrule(lr){21-23} \cmidrule(lr){24-26}

   & n\_imgs &
\multicolumn{1}{c}{\scriptsize \alg} & \multicolumn{1}{c}{\scriptsize GLOMAP} & \multicolumn{1}{c}{\scriptsize COLMAP} &
\multicolumn{1}{c}{\scriptsize \alg} & \multicolumn{1}{c}{\scriptsize GLOMAP} & \multicolumn{1}{c}{\scriptsize COLMAP} &
\multicolumn{1}{c}{\scriptsize \alg} & \multicolumn{1}{c}{\scriptsize GLOMAP} & \multicolumn{1}{c}{\scriptsize COLMAP} &
\multicolumn{1}{c}{\scriptsize \alg} & \multicolumn{1}{c}{\scriptsize GLOMAP} & \multicolumn{1}{c}{\scriptsize COLMAP} &
\multicolumn{1}{c}{\scriptsize \alg} & \multicolumn{1}{c}{\scriptsize GLOMAP} & \multicolumn{1}{c}{\scriptsize COLMAP} &
\multicolumn{1}{c}{\scriptsize \alg} & \multicolumn{1}{c}{\scriptsize GLOMAP} & \multicolumn{1}{c}{\scriptsize COLMAP} &
\multicolumn{1}{c}{\scriptsize \alg} & \multicolumn{1}{c}{\scriptsize GLOMAP} & \multicolumn{1}{c}{\scriptsize COLMAP} &
\multicolumn{1}{c}{\scriptsize \alg} & \multicolumn{1}{c}{\scriptsize GLOMAP} & \multicolumn{1}{c}{\scriptsize COLMAP} \\
\bottomrule
\end{tabular}

}
\caption{Per scene camera pose metrics on the NeRF-OSR Dataset.}
\label{tab:supp_pose_nosr}
\end{table*}

\begin{table*}[]
\resizebox{1.\linewidth}{!}{
    \tablestyle{2pt}{1.1}
\begin{tabular}{l r  r@{~}r@{~}r | r@{~}r@{~}r | r@{~}r@{~}r | r@{~}r@{~}r | r@{~}r@{~}r | r@{~}r@{~}r | r@{~}r@{~}r | r@{~}r@{~}r }

\toprule

  &  & \multicolumn{3}{c}{time (sec)} &
\multicolumn{3}{c}{ATE$\downarrow$} &
\multicolumn{3}{c}{RTA@3$\uparrow$} &
\multicolumn{3}{c}{RRA@3$\uparrow$} &
\multicolumn{3}{c}{RTA@1$\uparrow$} &
\multicolumn{3}{c}{RRA@1$\uparrow$} &
\multicolumn{3}{c}{AUC-R\&T @ 3 $\uparrow$} &
\multicolumn{3}{c}{AUC-R\&T @ 1 $\uparrow$} \\

\cmidrule(lr){3-5} \cmidrule(lr){6-8} \cmidrule(lr){9-11}
\cmidrule(lr){12-14} \cmidrule(lr){15-17} \cmidrule(lr){18-20}
\cmidrule(lr){21-23} \cmidrule(lr){24-26}

   & n\_imgs &
\multicolumn{1}{c}{\scriptsize \alg} & \multicolumn{1}{c}{\scriptsize GLOMAP} & \multicolumn{1}{c}{\scriptsize COLMAP} &
\multicolumn{1}{c}{\scriptsize \alg} & \multicolumn{1}{c}{\scriptsize GLOMAP} & \multicolumn{1}{c}{\scriptsize COLMAP} &
\multicolumn{1}{c}{\scriptsize \alg} & \multicolumn{1}{c}{\scriptsize GLOMAP} & \multicolumn{1}{c}{\scriptsize COLMAP} &
\multicolumn{1}{c}{\scriptsize \alg} & \multicolumn{1}{c}{\scriptsize GLOMAP} & \multicolumn{1}{c}{\scriptsize COLMAP} &
\multicolumn{1}{c}{\scriptsize \alg} & \multicolumn{1}{c}{\scriptsize GLOMAP} & \multicolumn{1}{c}{\scriptsize COLMAP} &
\multicolumn{1}{c}{\scriptsize \alg} & \multicolumn{1}{c}{\scriptsize GLOMAP} & \multicolumn{1}{c}{\scriptsize COLMAP} &
\multicolumn{1}{c}{\scriptsize \alg} & \multicolumn{1}{c}{\scriptsize GLOMAP} & \multicolumn{1}{c}{\scriptsize COLMAP} &
\multicolumn{1}{c}{\scriptsize \alg} & \multicolumn{1}{c}{\scriptsize GLOMAP} & \multicolumn{1}{c}{\scriptsize COLMAP} \\
\bottomrule
\end{tabular}

}
\caption{Per scene camera pose metrics on the DroneDeploy Dataset.}
\label{tab:supp_pose_dploy}
\end{table*}

\begin{table*}[]
\resizebox{1.\linewidth}{!}{
    \tablestyle{2pt}{1.1}
\begin{tabular}{l r  r@{~}r@{~}r | r@{~}r@{~}r | r@{~}r@{~}r | r@{~}r@{~}r | r@{~}r@{~}r | r@{~}r@{~}r | r@{~}r@{~}r | r@{~}r@{~}r }

\toprule

  &  & \multicolumn{3}{c}{time (sec)} &
\multicolumn{3}{c}{ATE$\downarrow$} &
\multicolumn{3}{c}{RTA@3$\uparrow$} &
\multicolumn{3}{c}{RRA@3$\uparrow$} &
\multicolumn{3}{c}{RTA@1$\uparrow$} &
\multicolumn{3}{c}{RRA@1$\uparrow$} &
\multicolumn{3}{c}{AUC-R\&T @ 3 $\uparrow$} &
\multicolumn{3}{c}{AUC-R\&T @ 1 $\uparrow$} \\

\cmidrule(lr){3-5} \cmidrule(lr){6-8} \cmidrule(lr){9-11}
\cmidrule(lr){12-14} \cmidrule(lr){15-17} \cmidrule(lr){18-20}
\cmidrule(lr){21-23} \cmidrule(lr){24-26}

   & n\_imgs &
\multicolumn{1}{c}{\scriptsize \alg} & \multicolumn{1}{c}{\scriptsize GLOMAP} & \multicolumn{1}{c}{\scriptsize COLMAP} &
\multicolumn{1}{c}{\scriptsize \alg} & \multicolumn{1}{c}{\scriptsize GLOMAP} & \multicolumn{1}{c}{\scriptsize COLMAP} &
\multicolumn{1}{c}{\scriptsize \alg} & \multicolumn{1}{c}{\scriptsize GLOMAP} & \multicolumn{1}{c}{\scriptsize COLMAP} &
\multicolumn{1}{c}{\scriptsize \alg} & \multicolumn{1}{c}{\scriptsize GLOMAP} & \multicolumn{1}{c}{\scriptsize COLMAP} &
\multicolumn{1}{c}{\scriptsize \alg} & \multicolumn{1}{c}{\scriptsize GLOMAP} & \multicolumn{1}{c}{\scriptsize COLMAP} &
\multicolumn{1}{c}{\scriptsize \alg} & \multicolumn{1}{c}{\scriptsize GLOMAP} & \multicolumn{1}{c}{\scriptsize COLMAP} &
\multicolumn{1}{c}{\scriptsize \alg} & \multicolumn{1}{c}{\scriptsize GLOMAP} & \multicolumn{1}{c}{\scriptsize COLMAP} &
\multicolumn{1}{c}{\scriptsize \alg} & \multicolumn{1}{c}{\scriptsize GLOMAP} & \multicolumn{1}{c}{\scriptsize COLMAP} \\
\bottomrule
\end{tabular}

}
\caption{Per scene camera pose metrics on several large-scale datasets including ZipNeRF, Mill-19, Urbanscene3D and Eyeful Tower.}
\label{tab:supp_pose_large}
\end{table*}

\begin{table}[]
\resizebox{1.\linewidth}{!}{
    \tablestyle{4pt}{1.1}
\begin{tabular}{l l ccc | ccc }
\toprule
& & \multicolumn{3}{c}{Instant-NGP} & \multicolumn{3}{c}{Gaussian Splatting}  \\
\cmidrule(lr){3-5} \cmidrule(lr){6-8}
& &
\scriptsize \alg & \scriptsize GLOMAP & \scriptsize COLMAP &
\scriptsize \alg & \scriptsize GLOMAP & \scriptsize COLMAP \\
\midrule
\multirow{7}{*}{\rotatebox{90}{training}} & Barn &  \cellCD[18pt]{ffffff}{23.37} &  \cellCD[18pt]{b7d7a8}{23.68} &  \cellCD[18pt]{b7d7a8}{23.69} &  \cellCD[18pt]{ffffff}{26.17} &  \cellCD[18pt]{b7d7a8}{27.81} &  \cellCD[18pt]{b7d7a8}{27.79} \\
 & Caterpillar &  \cellCD[18pt]{b7d7a8}{20.17} &  \cellCD[18pt]{b7d7a8}{20.20} &  \cellCD[18pt]{b7d7a8}{20.23} &  \cellCD[18pt]{b7d7a8}{23.30} &  \cellCD[18pt]{b7d7a8}{23.47} &  \cellCD[18pt]{b7d7a8}{23.59} \\
 & Courthouse &  \cellCD[18pt]{b7d7a8}{19.96} &  \cellCD[18pt]{ffffff}{14.73} &  \cellCD[18pt]{69a84f}{20.27} &  \cellCD[18pt]{b7d7a8}{21.12} &  \cellCD[18pt]{ffffff}{12.23} &  \cellCD[18pt]{69a84f}{22.25} \\
 & Ignatius &  \cellCD[18pt]{b7d7a8}{18.11} &  \cellCD[18pt]{b7d7a8}{18.14} &  \cellCD[18pt]{b7d7a8}{18.26} &  \cellCD[18pt]{ffffff}{21.42} &  \cellCD[18pt]{b7d7a8}{22.04} &  \cellCD[18pt]{b7d7a8}{21.86} \\
 & Meetingroom &  \cellCD[18pt]{ffffff}{21.61} &  \cellCD[18pt]{b7d7a8}{22.59} &  \cellCD[18pt]{b7d7a8}{22.41} &  \cellCD[18pt]{ffffff}{23.71} &  \cellCD[18pt]{b7d7a8}{25.35} &  \cellCD[18pt]{b7d7a8}{25.17} \\
 & Truck &  \cellCD[18pt]{b7d7a8}{21.19} &  \cellCD[18pt]{ffffff}{16.86} &  \cellCD[18pt]{69a84f}{21.45} &  \cellCD[18pt]{b7d7a8}{23.58} &  \cellCD[18pt]{ffffff}{18.34} &  \cellCD[18pt]{69a84f}{24.50} \\
\bottomrule
\end{tabular}

}
\caption{
Per-scene novel view synthesis results on Tanks and Temples.
}
\label{tab:nerf_tnt_full}
\end{table}

\section{More Results}
Additional speed benchmarking with different hardware configurations is reported in Tab.~\ref{tab:detailed_timing}.
We show per-scene pose accuracy metrics for all datasets from \cref{tab:supp_pose_m360} to \cref{tab:supp_pose_large}. The per-scene NeRF and Gaussian Splatting evaluation on Tanks and Temples is shown in \cref{tab:nerf_tnt_full}.

\section{Technical Details}
\label{app:technical}

\subsection{Distortion Estimation}
\label{app:distortion}

\noindent\textbf{Hierarchical search} In order to accelerate the interval search in distortion estimation, which scales linearly with the number of candidates, we employ a hierarchical search strategy that iteratively shrinks the interval. At each level of the hierarchy, after finding the solution, we set the left and right candidates as the endpoints of the new interval for the next level. The solution at the last level is the final estimate.

\noindent\textbf{Multiple cameras} If the two images in a pair do not share intrinsics, but the distortion parameter of one of the images is known, we can use a similar 1D search method to determine the distortion parameter of the other image. With this, we can extend the distortion estimation algorithm to deal with multiple different cameras, each of which corresponds to a known subset of images.

We say that an image pair is \textit{ready} for a camera if either
\begin{enumerate}
    \item both images correspond to that camera; or
    \item only one of the images corresponds to that camera, but the distortion parameter of the other image is already estimated.
\end{enumerate}
We estimate the distortion parameters for these cameras one by one. Each time, among the cameras whose distortion have not been estimated, we pick the one with the largest number of ready image pairs. The distortion parameter for this camera is then estimated using these image pairs. To do that, the only modification to the original algorithm (originally for a single image pair) is that for each candidate of $\alpha$ we compute the average epipolar error over all the point pairs in all the ready image pairs. The next camera is picked likewise, until the distortion parameters for all the cameras are estimated.

\begin{figure}[!t] %
    \centering
    \includegraphics[width=1.0\columnwidth]{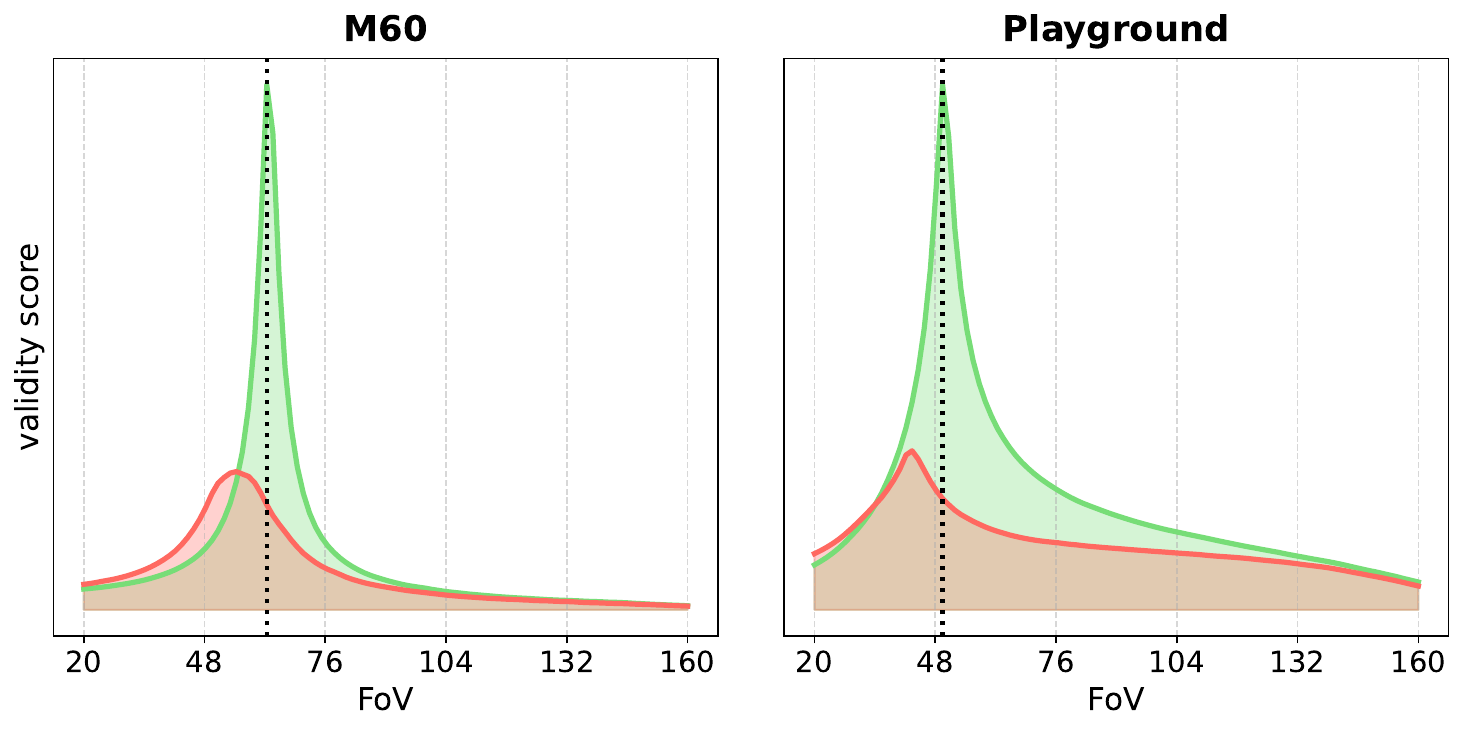}
    \caption{Effect of distortion on focal length estimation. Curves in  {\color{green}green} are with undistortion, and curves in {\color{red}red} without. The dotted lines indicate the ground-truth FoVs. %
    }
    \label{fig:ablation-distortion}
\end{figure}

\noindent\textbf{Importance of undistortion} The most direct impact of incorrect distortion is on focal length estimation as illustrated in Figure~\ref{fig:ablation-distortion}, which visualizes focal length validity (smoothed over discrete samples) on two of the scenes with and without distortion estimation. After keypoints are undistorted (green), the validity score peaks at an accurate FoV estimation. Without distortion estimation (red), the total validity score decreases drastically, and the peak deviates from the correct FoV.

\subsection{Focal Length Estimation}
\label{app:focal}

We adopt a similar strategy as distortion estimation~\cref{app:distortion} to deal with multiple cameras. However we do not use hierarchical sampling because processing each candidate is relatively cheap, and we can afford to densely sample the interval.

\subsection{Global Rotation}
\label{app:rotation}
\subsubsection{Initialization} \label{app:rotation-initialization}
In this section we discuss how to initialize the global rotation matrices for optimization. It is a modified version of \citet{martinec2007robust}.

Denote the set of images as $\mathcal{I}$ and the set of image pairs as $\mathcal{P}=\{(i, j)\}_{(i,j)\in\mathcal{I}\times\mathcal{I}}$, where each element has relative rotation matrix $\mR^{i\rightarrow j}$. The goal is to construct a solution of global world to camera rotation matrices $\{\mR_i\}_{i\in\mathcal{I}}$ to minimize the objective (note that in contrast to the final objective used in iterative optimization, this one uses L2 loss)

\begin{equation} \label{eq:appendix-rotation-objective}
    \mathcal{L} = \sum_{(i, j)\in\mathcal{P}}\left\Vert \mR^{(j)} - \mR^{i\rightarrow j} \mR^{(i)} \right\Vert^2
\end{equation}

Unfortunately, this problem cannot be directly solved using simple least square techniques since $\mR^{(i)}\in\text{SO}(3)$ are $3\times 3$ matrices with orthogonality constraints. However, we can decompose it into several sub-problems to circumvent the constraints. In the following we use $\mA_{*,k}$ to denote the $k^\text{th}$ solumn of a matrix $\mA$. Note that each term inside the summation in \cref{eq:appendix-rotation-objective} can be splitted into three parts

\begin{align}
    \mathcal{L} = \sum_{(i, j)\in\mathcal{P}} \sum_{k=1,2,3} \left\Vert \mR^{(j)}_{*,k} - \mR^{i\rightarrow j} \mR^{(i)}_{*,k} \right\Vert^2
\end{align}
Since $\mR^{(i)}$ is an orthogonal matrix, the column vectors $\mR^{(i)}_{*,k}$ have unit length and are mutually orthogonal. These orthogonality constraints are difficult to deal with, but if we only look at one particular column, say $k=1$, and ignore the unit length constraint, the objective becomes an unconstrained least squares problem
\begin{align}
    \mathcal{L}^{(1)} = \sum_{(i, j)\in\mathcal{P}}  \left\Vert \mR^{(j)}_{*,1} - \mR^{i\rightarrow j} \mR^{(i)}_{*,1} \right\Vert^2
\end{align}
Using techniques like SVD we can find a non-trivial solution that is not all zero. And since the relative rotation matrices $\mR^{i\rightarrow j}$ are also orthogonal matrices, left multiplying it with a vector preserves the vector length. This means that in the solution, the three-dimensional vectors $\mR^{(i)}_{*,1}$ should have similar lengths. We can normalize them to unit length and get a solution of the actual first columns of global rotation matrices $\{\mR^{(i)}\}_{i\in\mathcal{I}}$.

Given already estimated first columns  $\{\mR^{(i)}_{*,1}\}_{i\in\mathcal{I}}$ of the global rotation matrices, we can estimate the second columns by minimizing the same objective but adding additional constraints enforcing orthogonality to the first columns
\begin{equation}
    \begin{split}
        \mathcal{L}^{(2)} = &\frac{1}{|\mathcal{P}|} \sum_{(i, j)\in\mathcal{P}}  \left\Vert \mR^{(j)}_{*,2} - \mR^{i\rightarrow j} \mR^{(i)}_{*,2} \right\Vert^2   \\
        &+ \frac{1}{|\mathcal{I}|} \sum_{i\in\mathcal{I}} \left\Vert {\mR^{(i)}_{*,1}}^\top  \mR^{(i)}_{*,2} \right\Vert^2,
    \end{split}
\end{equation}
where $|\mathcal{P}|$ is the number of image pairs and $|\mathcal{I}|$ is the number of images, and they are used heuristically for controlling the relative weighting of the two terms. Note that in the above $\mR^{(i)}_{*,1}$ are already fixed and the only free variables are $\{\mR^{(i)}_{*,2}\}$. So this is still a least squares problem, and can be solved by applying SVD and then normalizing each 3-dimensional vector to unit length. A simple Gram-Schmidt process can be used for enforcing the orthogonality between the first and second columns.

We do not need to solve a least square problem again for the third columns because it can be directly computed by taking the cross product of the first two columns
\begin{align}
     \mR^{(i)}_{*,3} =  \mR^{(i)}_{*,1} \times  \mR^{(i)}_{*,2}
\end{align}

\subsubsection{Filtering}
To improve the robustness of global rotation alignment, we filter out some image pairs whose number of inlier point pairs does not exceed certain threshold. Determining this threshold can be tricky: a low threshold might introduce a lot of outlier image pairs, but a high threshold could reduce the number of connections and even disconnect the images into several clusters. To alleviate this problem, we start from a large threshold, and reduce it by half if it leads to disconnected clusters. We do this iteratively until either all the images are connected, or a minimal threshold is reached. This partially solves the problem by making the threshold adaptive to the data. However it is still common that even at the minimal threshold, a few images are disconnected from the others. In that case we just consider them as outliers and ignore them in rotation alignment and later stages.

\subsection{Tracks}
\label{app:track}

Consider the graph in which 2D keypoints are nodes and pairwise edges denote keypoint matches. When a 3D scene point is observed in $m$ different images, the projected 2D keypoints should ideally form a complete subgraph with $m$ vertices. 
In practice, keypoint matching has low recall and tends to miss many point-pairs. A subgraph of related keypoints is often far from fully connected. 
Since the number (and quality) of matches is critical to the accuracy of pose estimation, we 
make up for
low matching recall by \emph{track completion}.  A \emph{track} is a connected component 
in the
2D keypoint connectivity graph
and implies the existence of a shared 3D point. Tracks are used heavily in SfM~\cite{schoenberger2016sfm, pan2024glomap} to impose extra constraints, e.g., bundle adjustment~\cite{triggs2000bundle} initializes 3D points based on tracks and minimizes the re-projection error of each 3D point with its track members.

\alg avoids bundle adjustment and data structures containing both 3D scene points and 2D keypoints. Instead we explicitly convert tracks to additional matches 
with pairwise combinations of all keypoints in each track. 
This way we still make use of the transitivity of matching and benefit from the extra constraints. 
These additional point pairs 
are only introduced after global rotation alignment, and are used in the global translation alignment and epipolar adjustment steps described below.

\subsection{Global Translation}
\label{app:translation}

\subsubsection{Relative Translation}
Global translation alignment in our method relies heavily on relative translations between image pairs. Rather than using the translations determined via pose decomposition, we first re-estimate the relative translations. %
We do so for two reasons. First, since we have estimated the global rotations, we can go back and re-compute the relative rotation of any image pair. The relative rotation computed in this way is much more accurate than those from relative pose decomposition. In turn, a better estimate of the relative rotation enables us to more accurately estimate relative translation. Second, after generating new point pairs from tracks, some image pairs that originally had no matches might have some now, and we can use these new point pairs to estimate the relative translation. We use 2D grid search to re-estimate the unit relative translation vectors. We first sample a set of candidates on the surface of the unit sphere, evaluate the mean epipolar error of each sample, and choose the candidate with the lowest error as the final estimate.

\subsubsection{Multiple Initializations} 
\label{app:translation-initializations}

Although random initialization works surprisingly well for the objective in \cref{eq:global-translation-objective}, it occasionally produces a small number of outliers. 
To deal with the problem, we propose to do multiple independent runs from different random initializations, and merge the solutions as the initialization for the final optimization loop. Since the global rotations are the same, different solutions can be aligned by moving the centroid to the origin and rescaling uniformly to have unit average norm. Then for each image the solution with the lowest average loss is chosen to be in the merged result.

\subsection{Epipolar Adjustment}
\label{app:epipolar}

In this section we derive the following equivalent form of the L2 epipolar adjustment objective (the re-weighting objective is similar)
\begin{equation} \label{eq:appendix-epipolar-objective-l2}
\mathcal{L}\hspace{-.2em}=\hspace{-.2em} \frac{1}{Z}\sum_{n=1}^{|\mathcal{P}|}\sum_{m=1}^{|\tilde{\mathcal{Q}}_n|} ( \tilde{\vx}^{(2)\top}_{nm}\mE_n\tilde{\vx}^{(1)}_{nm} )^2=\frac{2}{Z}\sum_{n=1}^{|\mathcal{P}|} \ve_n^\top \mW_n \ve_n
\end{equation}

Note that each error term is linear in the essential matrix, so we can re-write it as a dot product of the a weight vector and a flattened version of the essential matrix
\begin{subequations}
    \begin{align}
        \mathcal{L} &= \frac{1}{Z}\sum_{n=1}^{|\mathcal{P}|}\sum_{m=1}^{|\tilde{\mathcal{Q}}_n|} ( \tilde{\vx}^{(2)\top}_{nm}\mE_{n}\tilde{\vx}^{(1)}_{nm} )^2 \\
        &= \frac{1}{Z}\sum_{n=1}^{|\mathcal{P}|}\sum_{m=1}^{|\tilde{\mathcal{Q}}_n|} ( (\tilde{\vx}^{(2)}_{nm}\tilde{\vx}^{(1)\top}_{nm}) \otimes \mE_{n} )^2 \\
        &= \frac{1}{Z}\sum_{n=1}^{|\mathcal{P}|}\sum_{m=1}^{|\tilde{\mathcal{Q}}_n|} ( \vw^{\top}_{nm}\ve_{n})^2,
    \end{align}
\end{subequations}
where $\otimes$ is the element-wise multiplication operator, $\vw_{nm}=\text{flatten}(\tilde{\vx}^{(2)}_{nm}\tilde{\vx}^{(1)\top}_{nm})\in\mathbb{R}^9$, and $\ve_{n}=\text{flatten}(\mE_{n})\in\mathbb{R}^9$.
Now re-arrange the terms to get the summation of a set of quadratic forms
\begin{subequations}
    \begin{align}
        \mathcal{L} 
        &= \frac{1}{Z}\sum_{n=1}^{|\mathcal{P}|}\sum_{m=1}^{|\tilde{\mathcal{Q}}_n|} ( \vw^{\top}_{nm}\ve_{n})^2 \\
        &= \frac{1}{Z}\sum_{n=1}^{|\mathcal{P}|}\sum_{m=1}^{|\tilde{\mathcal{Q}}_n|} ( \vw^{\top}_{nm}\ve_{n})^\top ( \vw^{\top}_{nm}\ve_{n}) \\
        &= \frac{2}{Z}\sum_{n=1}^{|\mathcal{P}|} \sum_{m=1}^{|\tilde{\mathcal{Q}}_n|}  \ve_{n}^\top \vw_{nm}\vw^{\top}_{nm}\ve_{n} \\
        &= \frac{2}{Z}\sum_{n=1}^{|\mathcal{P}|} \ve_{n}^\top \left(\sum_{m=1}^{|\tilde{\mathcal{Q}}_n|}  \vw_{nm}\vw^{\top}_{nm}\right)\ve_{n} \\
        &= \frac{2}{Z}\sum_{n=1}^{|\mathcal{P}|} \ve_{n}^\top \mW_n \ve_{n}
    \end{align}
\end{subequations}
where $\mW_n=\sum_{m=1}^{|\tilde{\mathcal{Q}}_n|}  \vw_{nm}\vw^{\top}_{nm}\in\mathbb{R}^{9\times9}$

\subsection{Sparse Reconstruction}
\label{app:sparse}

After pose estimation, we do a sparse reconstruction of the scene by triangulating the matched keypoint pairs from track completion. The 3D points corresponding to the same track are merged by averaging. To eliminate outliers, after merging the 3D points, we compute the re-projection error for each 2D keypoint, and mark those with large errors to be outlier keypoints. A 3D point is dropped if the number of inlier keypoints in the track is smaller than 3. We also filter out a 3D point if the maximal triangulation angle is smaller than some threshold.

\subsection{Data Ground Truth}
\label{app:data}

With the exception of Tanks and Temples, each of the datasets we evaluate on includes author-provided reference camera poses. These reference poses are obtained through different means, including COLMAP (for MipNeRF360, ZipNeRF, NeRF-OSR), PixSfM~\cite{lindenberger2021pixel} (for Mill-19 and Urbanscene3D), and commercial software (for DroneDeploy and Eyeful Tower). In the case of Urbanscene3D, we use the poses provided by \citet{turki2022mega}. For Tanks and Temples, we use COLMAP poses provided by \citet{kulhanek2024nerfbaselines}. On one of the scenes from Tanks and Temples (Courthouse), we found that the reference poses are obviously inconsistent with the images, but decided to still treat them as ground-truth.

\subsection{Additional Ablation Study}
\label{app:ablation}

\begin{table}[t]
    \centering
    \resizebox{1.0\linewidth}{!}{
    \tablestyle{8pt}{1.1}
\renewcommand{\arraystretch}{1.0}
\begin{tabular}{l l rrrrr}
\toprule
 &  &
\multicolumn{1}{c}{$m=1$} &
\multicolumn{1}{c}{$m=2$} &
\multicolumn{1}{c}{$m=3$} &
\multicolumn{1}{c}{$m=4$} &
\multicolumn{1}{c}{$m=8$} \\
\midrule
\multirow{2}{6em}{zipnerf\_nyc} & RTE@30 & \textbf{0.59} & \textbf{0.09} & 0.09 & 0.09 & 0.10 \\
                  & RTA@3 & 98.59 & 99.45 & 99.45 & 99.45 & 99.45 \\
\midrule
\multirow{2}{6em}{zipnerf\_alameda}  & RTE@30 & \textbf{0.32} & \textbf{0.14} & 0.13 & 0.09 & 0.09 \\
                  & RTA@3 & 97.90 & 98.28 & 98.29 & 98.40 & 98.39 \\
\midrule
\multirow{2}{6em}{tnt\_Train}        & RTE@30 & \textbf{1.40} & \textbf{0.02} & 0.02 & 0.02 & 0.29 \\
                  & RTA@3 & 97.15 & 99.62 & 99.64 & 99.61 & 99.08 \\
\midrule
\multirow{2}{6em}{tnt\_Lighthouse}   & RTE@30 & \textbf{1.98} & \textbf{0.01} & 0.01 & 0.02 & 0.01 \\
                  & RTA@3 & 96.82 & 99.26 & 99.23 & 99.34 & 99.33 \\
\midrule
\multirow{2}{6em}{dploy\_ruins3}     & RTE@30 & \textbf{1.23} & \textbf{0.66} & 0.67 & 0.92 & 0.69 \\
                  & RTA@3 & 92.70 & 94.56 & 94.44 & 94.06 & 93.77 \\
\midrule
\multirow{2}{6em}{dploy\_house4}     & RTE@30 & \textbf{3.82} & \textbf{0.83} & 1.00 & 0.83 & 0.83 \\
                  & RTA@3 & 93.00 & 96.92 & 95.69 & 97.48 & 95.48 \\
\bottomrule
\vspace{-5mm}
\end{tabular}

    }
    \caption{Ablation of multiple translation initializations on selected scenes. The results are obtained right after translation alignment and before epipolar adjustment. We bold the RTE@$30$ entries for $m=1$ and $m=2$ initializations to highlight its effect.}
    \label{tab:ablate_multiinit}
\end{table}

\begin{table}[t]
\centering
\resizebox{1.0\linewidth}{!}{
    \tablestyle{2pt}{1.1}
\renewcommand{\arraystretch}{1.0}
\scriptsize
\begin{tabular}{l @{~~~} l rrrrr }
\toprule
 & &
\multicolumn{1}{c}{AUC@3} &
\multicolumn{1}{c}{AUC@10} &
\multicolumn{1}{c}{RTA@1} &
\multicolumn{1}{c}{RTA@5} &
\multicolumn{1}{c}{RRA@3} \\
\midrule
\multirow{3}{4em}{m360 (9)} &  \alg & \textbf{97.2} & \textbf{99.1} & \textbf{99.8} & \textbf{100.0} & \textbf{100.0} \\
         &  w/o epipolar adjustment & 75.0 & 90.8 & 85.5 & 99.5 & 94.5 \\
         &  w/o track completion & 80.4 & 86.4 & 83.3 & 91.4 & 83.6 \\
\midrule
 \multirow{3}{4em}{alameda}  &  \alg & \textbf{95.2} & \textbf{98.1} & \textbf{99.0} & \textbf{99.3} & \textbf{99.9} \\
         &  w/o epipolar adjustment & 86.7 & 95.5 & 94.9 & 99.2 & 99.8 \\
         &  w/o track completion & 94.8 & \textbf{98.1} & \textbf{99.0} & 99.4 & \textbf{99.9} \\
\midrule
\multirow{3}{4em}{berlin}   &  \alg & \textbf{81.6} & \textbf{93.2} & \textbf{92.8} & \textbf{99.2} & \textbf{97.5} \\
         &  w/o epipolar adjustment & 70.4 & 89.5 & 82.4 & 98.8 & 95.7 \\
         &  w/o track completion & 60.4 & 81.3 & 70.4 & 90.8 & 90.3 \\
\midrule
\multirow{3}{4em}{london}   &  \alg & \textbf{96.6} & \textbf{98.8} & \textbf{99.6} & \textbf{99.9} & 99.7 \\
         &  w/o epipolar adjustment & 90.1 & 96.6 & 97.8 & 99.7 & 99.3 \\
         &  w/o track completion & 96.1 & 98.7 & \textbf{99.6} & \textbf{99.9} & \textbf{99.8} \\
\midrule
\multirow{3}{4em}{nyc}      &  \alg & \textbf{94.6} & \textbf{98.1} & 99.2 & 99.6 & 99.6 \\
         &  w/o epipolar adjustment & 89.7 & 96.7 & 97.0 & 99.7 & 99.6 \\
         &  w/o track completion & 93.9 & 98.0 & \textbf{99.4} & \textbf{99.8} &  \textbf{99.8} \\
\bottomrule
\end{tabular}

}
\caption{Epipolar adjustment and track completion ablation on the MipNeRF360~\cite{barron2022mipnerf360} and ZipNeRF~\cite{barron2023zipnerf} datasets. Results for MipNeRF360 are averaged over all the scenes, and for ZipNeRF each scene is listed separately.}
\label{tab:ablate_epioptim_and_tracks}
\end{table}

\subsubsection{Track completion} 
In Table~\ref{tab:ablate_epioptim_and_tracks} we show the final performance of the \alg with and without augmented point pairs from track completion (\cref{app:track}). Track completion significantly improves performance for  MipNeRF360 scenes and some but not all ZipNeRF scenes.

\subsubsection{Multiple translation initializations}
As shown in Table~\ref{tab:ablate_multiinit}, 
while a single initialization is prone to large-error outliers (see RTE@$30$ defined as $100.0$ - RTA@$30$), increasing the number of initalizations improves performance. However, the effect plateaus with increased initializations %
and does not completely fix the outlier problem.

\subsubsection{Epipolar adjustment} Table~\ref{tab:ablate_epioptim_and_tracks} also presents the performance of \alg with and without the final epipolar adjustment step. On all metrics, epipolar adjustment consistently improves  over the poses from global translation alignment. The improvement is more prominent for stricter metrics (RTA@1), but less so for more tolerant metrics like RTA@5. This suggests that after translation alignment the cameras are already roughly in place, and epipolar adjustment continues to squeeze as much precision as it can.

\begin{table}[]
\resizebox{1.\linewidth}{!}{
    
\tablestyle{2pt}{1.1}
\begin{tabular}{l r  r@{~}r | r@{~}r | r@{~}r | r@{~}r r }
\toprule
  &  &
\multicolumn{2}{c}{ATE$\downarrow$} &
\multicolumn{2}{c}{RTA@3$\uparrow$} &
\multicolumn{2}{c}{AUC-R\&T @ 3 $\uparrow$} &
\multicolumn{2}{c}{AUC-R\&T @ 1 $\uparrow$} \\
\cmidrule(lr){3-4} \cmidrule(lr){5-6} \cmidrule(lr){7-8} \cmidrule(lr){9-10}
   & n\_imgs &
\multicolumn{1}{c}{\scriptsize \alg} & \multicolumn{1}{c}{\scriptsize GLOMAP} &
\multicolumn{1}{c}{\scriptsize \alg} & \multicolumn{1}{c}{\scriptsize GLOMAP} &
\multicolumn{1}{c}{\scriptsize \alg} & \multicolumn{1}{c}{\scriptsize GLOMAP} &
\multicolumn{1}{c}{\scriptsize \alg} & \multicolumn{1}{c}{\scriptsize GLOMAP} \\
\midrule
botanical\_garden &   30 &  \cellCD[22pt]{b7d7a8}{8.2e-3} &  \cellCD[22pt]{69a84f}{4.3e-4} &  \cellCD[22pt]{b7d7a8}{86.9} &  \cellCD[22pt]{69a84f}{100.0} &  \cellCD[22pt]{b7d7a8}{68.3} &  \cellCD[22pt]{69a84f}{94.3} &  \cellCD[22pt]{b7d7a8}{52.0} &  \cellCD[22pt]{69a84f}{83.8} \\
       boulders &   26 &  \cellCD[22pt]{b7d7a8}{6.7e-4} &  \cellCD[22pt]{69a84f}{1.4e-4} &  \cellCD[22pt]{b7d7a8}{99.1} &  \cellCD[22pt]{b7d7a8}{100.0} &  \cellCD[22pt]{b7d7a8}{91.2} &  \cellCD[22pt]{69a84f}{97.0} &  \cellCD[22pt]{b7d7a8}{76.2} &  \cellCD[22pt]{69a84f}{91.0} \\
         bridge &  110 &  \cellCD[22pt]{b7d7a8}{1.3e-2} &  \cellCD[22pt]{69a84f}{2.0e-5} &  \cellCD[22pt]{b7d7a8}{92.9} &  \cellCD[22pt]{69a84f}{100.0} &  \cellCD[22pt]{b7d7a8}{85.3} &  \cellCD[22pt]{69a84f}{97.7} &  \cellCD[22pt]{b7d7a8}{73.3} &  \cellCD[22pt]{69a84f}{93.1} \\
      courtyard &   38 &  \cellCD[22pt]{b7d7a8}{3.8e-2} &  \cellCD[22pt]{69a84f}{1.8e-4} &  \cellCD[22pt]{f4cccc}{18.9} &  \cellCD[22pt]{69a84f}{100.0} &  \cellCD[22pt]{f4cccc}{6.9} &  \cellCD[22pt]{69a84f}{96.0} &  \cellCD[22pt]{f4cccc}{2.2} &  \cellCD[22pt]{69a84f}{88.2} \\
 delivery\_area &   44 &  \cellCD[22pt]{b7d7a8}{8.4e-2} &  \cellCD[22pt]{69a84f}{8.1e-5} &  \cellCD[22pt]{f4cccc}{23.6} &  \cellCD[22pt]{69a84f}{100.0} &  \cellCD[22pt]{f4cccc}{13.9} &  \cellCD[22pt]{69a84f}{97.8} &  \cellCD[22pt]{f4cccc}{6.1} &  \cellCD[22pt]{69a84f}{93.3} \\
           door &    7 &  \cellCD[22pt]{f4cccc}{-\vphantom{1.0}} &  \cellCD[22pt]{69a84f}{1.2e-4} &  \cellCD[22pt]{f4cccc}{-\vphantom{1.0}} &  \cellCD[22pt]{69a84f}{100.0} &  \cellCD[22pt]{f4cccc}{-\vphantom{1.0}} &  \cellCD[22pt]{69a84f}{98.0} &  \cellCD[22pt]{f4cccc}{-\vphantom{1.0}} &  \cellCD[22pt]{69a84f}{94.1} \\
        electro &   45 &  \cellCD[22pt]{b7d7a8}{7.5e-2} &  \cellCD[22pt]{69a84f}{3.0e-2} &  \cellCD[22pt]{b7d7a8}{86.3} &  \cellCD[22pt]{69a84f}{95.2} &  \cellCD[22pt]{b7d7a8}{76.9} &  \cellCD[22pt]{69a84f}{91.1} &  \cellCD[22pt]{b7d7a8}{61.6} &  \cellCD[22pt]{69a84f}{84.1} \\
exhibition\_hall &   68 &  \cellCD[22pt]{b7d7a8}{7.0e-2} &  \cellCD[22pt]{b7d7a8}{6.9e-2} &  \cellCD[22pt]{f4cccc}{2.8} &  \cellCD[22pt]{69a84f}{45.1} &  \cellCD[22pt]{f4cccc}{0.9} &  \cellCD[22pt]{69a84f}{40.9} &  \cellCD[22pt]{f4cccc}{0.1} &  \cellCD[22pt]{f4cccc}{34.3} \\
         facade &   76 &  \cellCD[22pt]{b7d7a8}{6.5e-2} &  \cellCD[22pt]{69a84f}{9.7e-5} &  \cellCD[22pt]{b7d7a8}{71.0} &  \cellCD[22pt]{69a84f}{100.0} &  \cellCD[22pt]{b7d7a8}{66.8} &  \cellCD[22pt]{69a84f}{97.4} &  \cellCD[22pt]{b7d7a8}{60.8} &  \cellCD[22pt]{69a84f}{92.4} \\
         kicker &   31 &  \cellCD[22pt]{69a84f}{5.9e-4} &  \cellCD[22pt]{b7d7a8}{1.6e-2} &  \cellCD[22pt]{69a84f}{98.5} &  \cellCD[22pt]{b7d7a8}{93.8} &  \cellCD[22pt]{b7d7a8}{86.6} &  \cellCD[22pt]{69a84f}{91.7} &  \cellCD[22pt]{b7d7a8}{65.0} &  \cellCD[22pt]{69a84f}{88.1} \\
  lecture\_room &   23 &  \cellCD[22pt]{b7d7a8}{3.0e-2} &  \cellCD[22pt]{69a84f}{2.5e-4} &  \cellCD[22pt]{b7d7a8}{84.2} &  \cellCD[22pt]{69a84f}{100.0} &  \cellCD[22pt]{b7d7a8}{71.7} &  \cellCD[22pt]{69a84f}{95.0} &  \cellCD[22pt]{b7d7a8}{55.3} &  \cellCD[22pt]{69a84f}{85.7} \\
   living\_room &   65 &  \cellCD[22pt]{b7d7a8}{1.3e-4} &  \cellCD[22pt]{69a84f}{8.4e-5} &  \cellCD[22pt]{b7d7a8}{99.7} &  \cellCD[22pt]{b7d7a8}{99.8} &  \cellCD[22pt]{b7d7a8}{95.3} &  \cellCD[22pt]{b7d7a8}{96.2} &  \cellCD[22pt]{b7d7a8}{86.8} &  \cellCD[22pt]{69a84f}{89.2} \\
         lounge &   10 &  \cellCD[22pt]{b7d7a8}{9.6e-2} &  \cellCD[22pt]{b7d7a8}{9.5e-2} &  \cellCD[22pt]{f4cccc}{33.3} &  \cellCD[22pt]{f4cccc}{33.3} &  \cellCD[22pt]{f4cccc}{32.3} &  \cellCD[22pt]{f4cccc}{32.7} &  \cellCD[22pt]{f4cccc}{30.2} &  \cellCD[22pt]{f4cccc}{31.4} \\
         meadow &   15 &  \cellCD[22pt]{b7d7a8}{1.4e-1} &  \cellCD[22pt]{b7d7a8}{1.4e-1} &  \cellCD[22pt]{f4cccc}{13.3} &  \cellCD[22pt]{69a84f}{86.7} &  \cellCD[22pt]{f4cccc}{7.7} &  \cellCD[22pt]{69a84f}{80.2} &  \cellCD[22pt]{f4cccc}{4.9} &  \cellCD[22pt]{69a84f}{68.2} \\
    observatory &   27 &  \cellCD[22pt]{b7d7a8}{6.5e-3} &  \cellCD[22pt]{69a84f}{5.8e-4} &  \cellCD[22pt]{b7d7a8}{94.9} &  \cellCD[22pt]{69a84f}{99.1} &  \cellCD[22pt]{b7d7a8}{76.5} &  \cellCD[22pt]{69a84f}{86.5} &  \cellCD[22pt]{b7d7a8}{48.5} &  \cellCD[22pt]{69a84f}{63.9} \\
         office &   26 &  \cellCD[22pt]{b7d7a8}{9.7e-3} &  \cellCD[22pt]{69a84f}{7.6e-4} &  \cellCD[22pt]{b7d7a8}{54.8} &  \cellCD[22pt]{69a84f}{95.7} &  \cellCD[22pt]{b7d7a8}{43.9} &  \cellCD[22pt]{69a84f}{82.7} &  \cellCD[22pt]{f4cccc}{34.5} &  \cellCD[22pt]{69a84f}{61.2} \\
  old\_computer &   54 &  \cellCD[22pt]{b7d7a8}{6.8e-2} &  \cellCD[22pt]{b7d7a8}{5.6e-2} &  \cellCD[22pt]{f4cccc}{21.7} &  \cellCD[22pt]{69a84f}{65.3} &  \cellCD[22pt]{f4cccc}{16.0} &  \cellCD[22pt]{69a84f}{60.9} &  \cellCD[22pt]{f4cccc}{9.8} &  \cellCD[22pt]{69a84f}{53.5} \\
          pipes &   14 &  \cellCD[22pt]{b7d7a8}{5.8e-4} &  \cellCD[22pt]{69a84f}{2.6e-4} &  \cellCD[22pt]{b7d7a8}{98.9} &  \cellCD[22pt]{b7d7a8}{100.0} &  \cellCD[22pt]{b7d7a8}{92.5} &  \cellCD[22pt]{69a84f}{97.4} &  \cellCD[22pt]{b7d7a8}{79.8} &  \cellCD[22pt]{69a84f}{92.3} \\
     playground &   38 &  \cellCD[22pt]{b7d7a8}{8.3e-4} &  \cellCD[22pt]{69a84f}{1.1e-4} &  \cellCD[22pt]{b7d7a8}{99.4} &  \cellCD[22pt]{b7d7a8}{99.9} &  \cellCD[22pt]{b7d7a8}{89.4} &  \cellCD[22pt]{69a84f}{97.1} &  \cellCD[22pt]{b7d7a8}{70.8} &  \cellCD[22pt]{69a84f}{91.7} \\
         relief &   31 &  \cellCD[22pt]{b7d7a8}{6.1e-3} &  \cellCD[22pt]{69a84f}{7.2e-5} &  \cellCD[22pt]{b7d7a8}{77.8} &  \cellCD[22pt]{69a84f}{100.0} &  \cellCD[22pt]{b7d7a8}{62.1} &  \cellCD[22pt]{69a84f}{98.4} &  \cellCD[22pt]{b7d7a8}{48.9} &  \cellCD[22pt]{69a84f}{95.2} \\
      relief\_2 &   31 &  \cellCD[22pt]{b7d7a8}{3.7e-4} &  \cellCD[22pt]{69a84f}{7.9e-5} &  \cellCD[22pt]{b7d7a8}{99.8} &  \cellCD[22pt]{b7d7a8}{100.0} &  \cellCD[22pt]{b7d7a8}{94.5} &  \cellCD[22pt]{69a84f}{98.4} &  \cellCD[22pt]{b7d7a8}{84.3} &  \cellCD[22pt]{69a84f}{95.1} \\
         statue &   11 &  \cellCD[22pt]{b7d7a8}{5.5e-5} &  \cellCD[22pt]{69a84f}{2.3e-5} &  \cellCD[22pt]{b7d7a8}{100.0} &  \cellCD[22pt]{b7d7a8}{100.0} &  \cellCD[22pt]{b7d7a8}{99.5} &  \cellCD[22pt]{b7d7a8}{99.7} &  \cellCD[22pt]{b7d7a8}{98.5} &  \cellCD[22pt]{b7d7a8}{99.0} \\
        terrace &   23 &  \cellCD[22pt]{b7d7a8}{2.1e-4} &  \cellCD[22pt]{69a84f}{1.2e-4} &  \cellCD[22pt]{b7d7a8}{100.0} &  \cellCD[22pt]{b7d7a8}{100.0} &  \cellCD[22pt]{b7d7a8}{97.5} &  \cellCD[22pt]{b7d7a8}{97.7} &  \cellCD[22pt]{b7d7a8}{92.5} &  \cellCD[22pt]{b7d7a8}{93.1} \\
     terrace\_2 &   13 &  \cellCD[22pt]{b7d7a8}{2.6e-4} &  \cellCD[22pt]{b7d7a8}{2.2e-4} &  \cellCD[22pt]{b7d7a8}{100.0} &  \cellCD[22pt]{b7d7a8}{100.0} &  \cellCD[22pt]{b7d7a8}{96.6} &  \cellCD[22pt]{b7d7a8}{96.9} &  \cellCD[22pt]{b7d7a8}{89.9} &  \cellCD[22pt]{b7d7a8}{90.8} \\
       terrains &   42 &  \cellCD[22pt]{b7d7a8}{1.3e-3} &  \cellCD[22pt]{69a84f}{2.1e-4} &  \cellCD[22pt]{b7d7a8}{94.4} &  \cellCD[22pt]{69a84f}{99.8} &  \cellCD[22pt]{b7d7a8}{70.9} &  \cellCD[22pt]{69a84f}{94.6} &  \cellCD[22pt]{b7d7a8}{39.3} &  \cellCD[22pt]{69a84f}{84.6} \\
\bottomrule
\end{tabular}

}
\caption{
Per scene camera pose metrics on ETH3D.
}
\label{tab:eth3d}
\end{table}

\begin{table}[]
\resizebox{1.\linewidth}{!}{
    \tablestyle{4pt}{1.1}
\begin{tabular}{l cc  cc  cc}
\toprule
 & \multicolumn{2}{c}{Recall@1m$\uparrow$} & \multicolumn{2}{c}{AUC@1m$\uparrow$} & \multicolumn{2}{c}{AUC@5m$\uparrow$} \\
\cmidrule(lr){2-3} \cmidrule(lr){4-5} \cmidrule(lr){6-7}
 & \multicolumn{1}{c}{\scriptsize \alg} & \multicolumn{1}{c}{\scriptsize GLOMAP} & \multicolumn{1}{c}{\scriptsize \alg} & \multicolumn{1}{c}{\scriptsize GLOMAP} & \multicolumn{1}{c}{\scriptsize \alg} & \multicolumn{1}{c}{\scriptsize GLOMAP} \\
\midrule 
CAB & 4.77 & 11.6 & 4.32 & 4.7 & 4.74 & 16.9 \\
HGE & 5.94 & 48.4 & 5.26 & 22.2 & 5.70 & 50.3 \\
LIN & 7.16 & 87.3 & 3.95 & 46.7 & 7.96 & 85.6 \\
\bottomrule
\end{tabular}

}
\caption{
Per scene camera pose metrics on LaMAR.
}
\label{tab:lamar}
\end{table}

\section{Limitations}
\label{app:limitations}

\noindent\textbf{Sparse Views.} Our method assumes that the input images densely cover the 3D scene. Many components in the pipeline implicitly assume that the coverage is dense so that the negative effect of outlier image or point pairs will be averaged out. If the coverage is sparse, the pipeline will be sensitive to outliers and likely break down. We tested our method on the ETH3D MVS (DSLR)~\cite{schops2017multi}, where each scene only contains a small number of images, and show the results in \cref{tab:eth3d}. While \alg still succeeds on many scenes, it is less robust than GLOMAP.

\noindent\textbf{Intrinsics Estimation.} The intrinsics estimation algorithms in our method can fail under certain cases. Since the interval search used in both distortion and focal length estimation requires at least one image pair of images with shared intrinsics to begin with, it will not work if all the images have different intrinsics. It is also not robust if the number of images for each distinct camera is small. In addition, the focal length extraction method relies entirely on fundamental matrices, and is unreliable when the scene is dominated by homographies.

\noindent\textbf{Homography.} Apart from the impact on focal length estimation, too many homography image pairs can also jeopardize relative pose decomposition. Both essential and homography decomposition produce four different solutions, and they are usually disambiguated with a cheirality check. But for some homography and keypoint pairs, cheirality check is not enough for determining a unique solution. Our current strategy is to simply pick the solution with the lowest index if there is a tie. This has potential issues if there are too many homography image pairs.

\noindent\textbf{Repetitive Patterns and Symmetric Structures} Non-learning based keypoint features and matching are not robust in the cases of repetitive patterns and symmetric structures in the scene. These wrong matches are hard to filter because they can have a lot of inlier point pairs with a very consistent two-view geometric model. Most traditional SfM methods are more or less impacted by these erroneous matches, and so is ours. In our experiments this problem is most prominent in the advanced split of the Tanks and Temples dataset (\cref{tab:supp_pose_tnt}).

\noindent\textbf{Degenerate Motions} One important reason why bundle adjustment is popular in previous SfM methods is that it can utilize 3D points to resolve some ambiguities in global translation estimation. For example, when all the cameras are aligned in the same line, optimization methods that solely rely on relative motions or epipolar errors might fail because there is no way to uniquely (up to scale) determine the distance of any pair of cameras. Bundle adjustment uses tracks to impose extra constraints to solve this problem. This scenario is commonly seen in SLAM-like datasets. We tested our method on the large-scale LaMAR~\cite{sarlin2022lamar} dataset and show the results in the \cref{tab:lamar}. Each scene in LaMAR consists of multiple trajectories of a moving VR headset or hand-held phone. These trajectories contain many straight and forward motions, and different trajectories only overlap sparsely. Our method does not work well compared to GLOMAP.

\end{document}